\documentclass[conference]{IEEEtran}
\IEEEoverridecommandlockouts
\usepackage{cite}
\usepackage{amsmath,amssymb,amsfonts}
\usepackage{algorithmic}
\usepackage{graphicx}
\usepackage{textcomp}
\usepackage{xcolor}
\def\BibTeX{{\rm B\kern-.05em{\sc i\kern-.025em b}\kern-.08em
    T\kern-.1667em\lower.7ex\hbox{E}\kern-.125emX}}

\usepackage{times}
\usepackage{soul}
\usepackage{url}
\usepackage[hidelinks]{hyperref}
\usepackage[utf8]{inputenc}
\usepackage[small]{caption}
\usepackage{graphicx}
\usepackage{amsmath}
\usepackage{amsthm}
\usepackage{booktabs}
\usepackage{algorithm}
\usepackage{graphicx}
\usepackage{multirow}
\usepackage{subfigure}
\usepackage{amssymb}
\usepackage{algorithmic}
\usepackage{graphics}
\usepackage{threeparttable} 
\definecolor{baselinecolor}{gray}{0.9}
\definecolor{deemph}{gray}{0.6}

\newcommand{\gc}[1]{\textcolor{deemph}{#1}}
\usepackage{graphicx}  %插入图片的宏包
\usepackage{float}  %设置图片浮动位置的宏包
\usepackage{subfigure}  %插入多图时用子图显示的宏包
\usepackage{colortbl}
\definecolor{pink}{rgb}{.99,.91,.95}
\usepackage{graphics}
\usepackage{multirow}
\usepackage{pdfpages}
\usepackage{amsmath, amssymb, amsthm, xspace, color}

\newcommand\figcaption{\def\@captype{figure}\caption}
\newcommand\tabcaption{\def\@captype{table}\caption}
\makeatother
\usepackage{color}
\definecolor{revision_color}{RGB}{46, 139, 87}
\definecolor{first_color}{RGB}{220, 220, 220}

\begin{document}

\title{Learning from Noisy Crowd Labels with Logics
\thanks{*Corresponding author.}}

\author{\IEEEauthorblockN{
Zhijun Chen$^{1,2}$, Hailong Sun$^{1,2*}$, Haoqian He$^{1,2}$, Pengpeng Chen$^{1,2,3}$}
\IEEEauthorblockA{
SKLSDE Lab, Beihang University, Beijing, China$^1$ \\
%SKLSDE Lab, School of Computer Science and Engineering, Beihang University, Beijing, China$^1$ \\
%SKLSDE Lab, School of Software, Beihang University, Beijing, China$^2$ \\
Beijing Advanced Innovation Center for Big Data and Brain Computing, Beihang University, Beijing, China$^2$ \\
China's Aviation System Engineering Research Institute, Beijing, China$^3$\\
\{zhijunchen, sunhl, hehaoqian,  chenpp\}@buaa.edu.cn\
}}

\maketitle

\begin{abstract}
This paper explores the integration of symbolic
logic
knowledge into deep neural networks for learning from noisy crowd labels. We introduce 
\emph{Logic-guided Learning from Noisy Crowd Labels} (Logic-LNCL), an EM-alike iterative logic knowledge distillation framework that learns from both noisy labeled data and logic rules of interest. Unlike traditional EM methods, our framework contains a  ``pseudo-E-step'' that distills from the logic rules a new type of learning target, which is then used in the ``pseudo-M-step'' for training the classifier. Extensive evaluations on two real-world datasets for text sentiment classification and named entity recognition demonstrate that the proposed framework improves the state-of-the-art and provides a new solution to learning from noisy crowd labels.
\end{abstract}

\begin{IEEEkeywords}
crowdsourcing, noisy labels, weak supervision, neural-symbolic learning
\end{IEEEkeywords}
\setcounter{section}{0}

\section{Introduction}
\label{introduction}

\begin{figure*}[!t]
\centering
\includegraphics[scale=0.52]{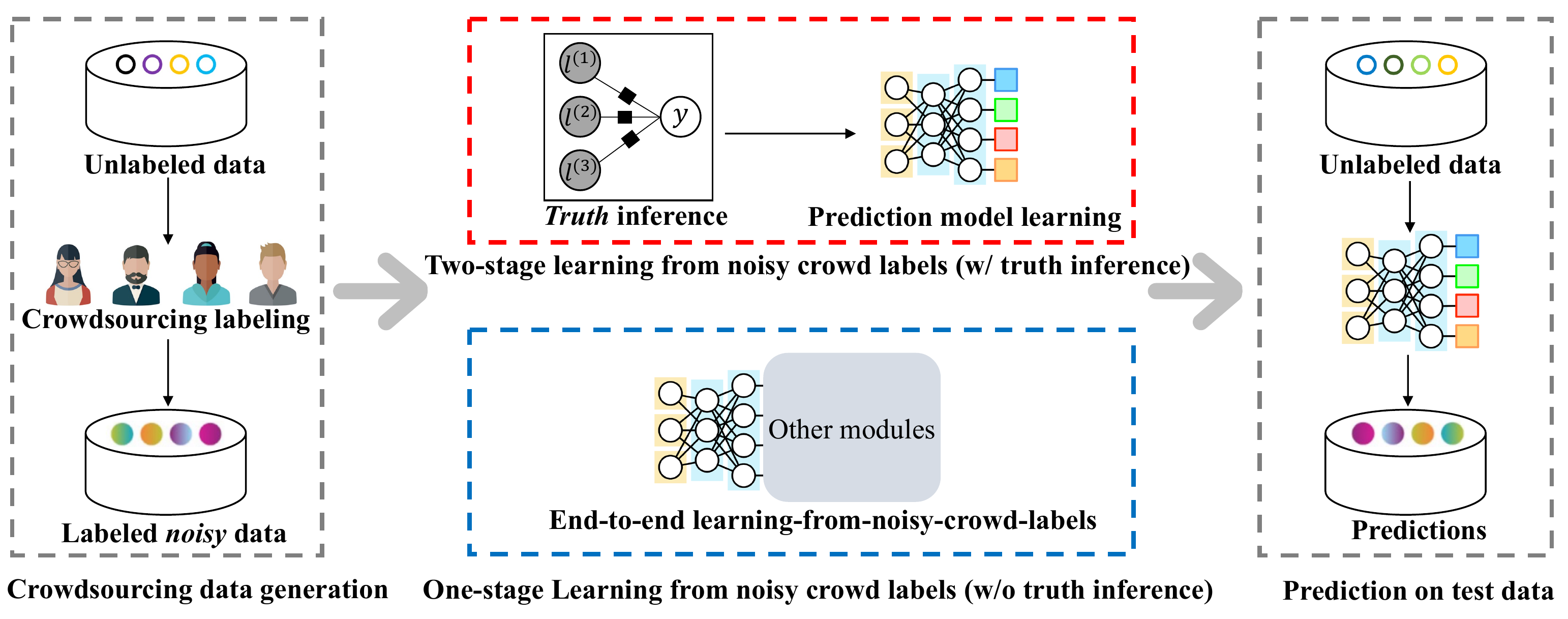} % Reduce the figure size so that it is slightly narrower than the column. Don't use precise values for figure width.This setup will avoid overfull boxes.
\caption{Overview of learning-from-noisy-crowd-labels pipeline.}
\label{figure:pipline}
\end{figure*}

Deep learning  heavily relies on the size and quality of labeled training data. Crowdsourced data annotation offers a time- and cost-efficient means to collect large-scale datasets for deep learning tasks~\cite{chai2019crowdsourcing}.
However, due to the imperfect crowd annotators, crowdsourced labels are often noisy; this has become a key concern for training deep neural networks~\cite{yang2019scalpel}.

To learn a classifier in the context of noisy crowd labels,  existing methods fall into two paradigms, as shown in Figure~\ref{figure:pipline}:
(1) 
The two-stage learning-from-noisy-crowds-labels (LNCL) paradigm that first uses 
\textit{truth inference} methods~\cite{dawid1979maximum,whitehill2009whose,welinder2010multidimensional,han2016incorporating, kim2012bayesian,simpson2018bayesian, nguyen2017aggregating,aydin2014crowdsourcing, li2014confidence} to infer  the true label of each instance and then performs traditional supervised learning;
(2) 
The one-stage LNCL paradigm that directly learns from noisy crowds labels in an end-to-end manner using ad hoc LNCL methods~\cite{raykar2010learning,rodrigues2013learning,rodrigues2014sequence, albarqouni2016aggnet,atarashi2018semi, rodrigues2018deep, guan2018said, cao2019max,chen2020structured,li2021learning}, which use all available information together to perform global optimization to obtain the classifier of interest and generally result in better performance.
Specifically, existing work on the recently prevailing one-stage LNCL mainly take two approaches: probabilistic or deep learning. 
The probabilistic approach models true labels as latent variables and uses expectation maximization (EM) to infer the true labels and to learn the parameters of the classifier (which generally be a deep neural network) in an iterative manner~\cite{raykar2010learning,rodrigues2013learning,rodrigues2014sequence, albarqouni2016aggnet,atarashi2018semi}.
More recent efforts have focused on the deep learning approach, which trains a deep neural architecture from noisy labels in an end-to-end manner, by introducing into the  architecture an extra neural layer that maps latent true labels to the noisy crowd labels~\cite{rodrigues2018deep,cao2019max,chen2020structured,li2021learning}.

In the broader AI community, there is a well-known thorny problem for all deep neural networks:
while being flexible to capture complex mapping between the features and label, deep learning approaches are generally data-hungry and not robust---they only learn (possibly spurious) statistical correlations. 
In learning from noisy crowd labels,
all existing prevailing one-stage LNCL approaches suffer much more from those issues for at least the following two reasons.   
(1) On the one hand, those approaches contain additional parameters that model the reliability of \emph{each} individual crowd annotator; consequently, training those approaches---that contain parameters of both neural network classifiers and annotators---requires a much larger amount of data than the conventional neural networks; 
(2) On the other hand, the presence of label noise makes it harder for the neural network to recognize relevant information from the training data for classification. As a result, the learned data representations may turn out to be dominated by information irrelevant to the true labels, thus becoming useless or even negatively impacting the performance. 

In comparison to learning-based, data-driven approaches,  reasoning-based, knowledge-driven approaches are more sample-efficient and more robust 
because of their capability of representing concepts and the causal relations among them. 
Recent discussions in the AI communities have converged on the idea of integrating symbolic approaches with machine learning~\cite{bengio2019system,garcez2020neurosymbolic}. 
A visible trend  is the growing body of work on neural-symbolic methods. 
For example, Allamanis et al.~\cite{allamanis2017learning} propose to learn continuous representations of symbolic knowledge for integration into neural networks;
Xu et al.~\cite{xu2018semantic} present the semantic loss that augments the learning objective of neural networks with soft-constraints specified with domain knowledge.
On the application side, neural-symbolic methods have been applied to both vision and language tasks including
visual relation prediction~\cite{xie2019embedding},
visual question answering (VQA)~\cite{yi2018neural},
text sentiment analysis~\cite{hu2016harnessing},
and semantic parsing~\cite{yin2018structvae}.

In this paper, we move one step further and explore:
\emph{in the presence of noisy crowd labels that makes the model more vulnerable and data-hungry, how can we capitalize on the interpretable and robust logic knowledge in learning from noisy crowd labels to improve the model’s generalization?}
Despite the obvious benefit of the neural-symbolic paradigm, it remains unclear how to best exploit logic knowledge in our context---existing neural-symbolic methods are not easily applicable due to the specific need for the inference of true labels from noisy crowd labels.
Specifically, properly inferring true labels needs to take into account the influence of logic knowledge and the reliability of annotators, while learning the parameters of the classifiers in a unified and principled framework. 
To address these problem, we introduce 
\emph{Logic-guided Learning from Noisy Crowd Labels}
(Logic-LNCL), an EM-alike iterative logic knowledge distillation framework that allows  neural networks to learn simultaneously from the noisy labels and logic rules. 
Unlike all existing probabilistic  models for learning from crowd labels,
in our  ``pseudo-E-step'', we introduce a new type of learning target
by adapting the original learning target with posterior regularization defined by the rules.
Then, inspired by knowledge distillation~\cite{hinton2015distilling}, both the original and adapted learning targets are used in the  ``pseudo-M-step" to update the parameters of the deep neural networks and those of the crowd annotators. 
By doing so, our framework seamlessly integrates symbolic logic knowledge and the inference for true labels, while allowing the deep neural network to 
\textit{learn from both data and logic knowledge}.

\textbf{Contributions.}
To the best of our knowledge, this is the first work to 
incorporate logic rules in learning from noisy (crowd) labels problems with a principled framework. 
We introduce Logic-LNCL, an EM-alike iterative  logic knowledge distillation framework 
for learning from both noisy crowd labels and logic rules.
We demonstrate the versatility of Logic-LNCL by instantiating it on two representative language tasks,
text sentiment classification and named entity recognition.
We conduct extensive evaluations of our proposed framework on real-world datasets, showing that Logic-LNCL improves the classifier's generalization and outperforms the state-of-the-art.

\section{Preliminaries}
\label{preliminaries}

\subsection{Problem Formulation: Learning from Noisy Crowd Data}
\label{Problem Formulation: Learning from Noisy Data}

\begin{figure}[t]
\centering
\includegraphics[width=0.26\textwidth]{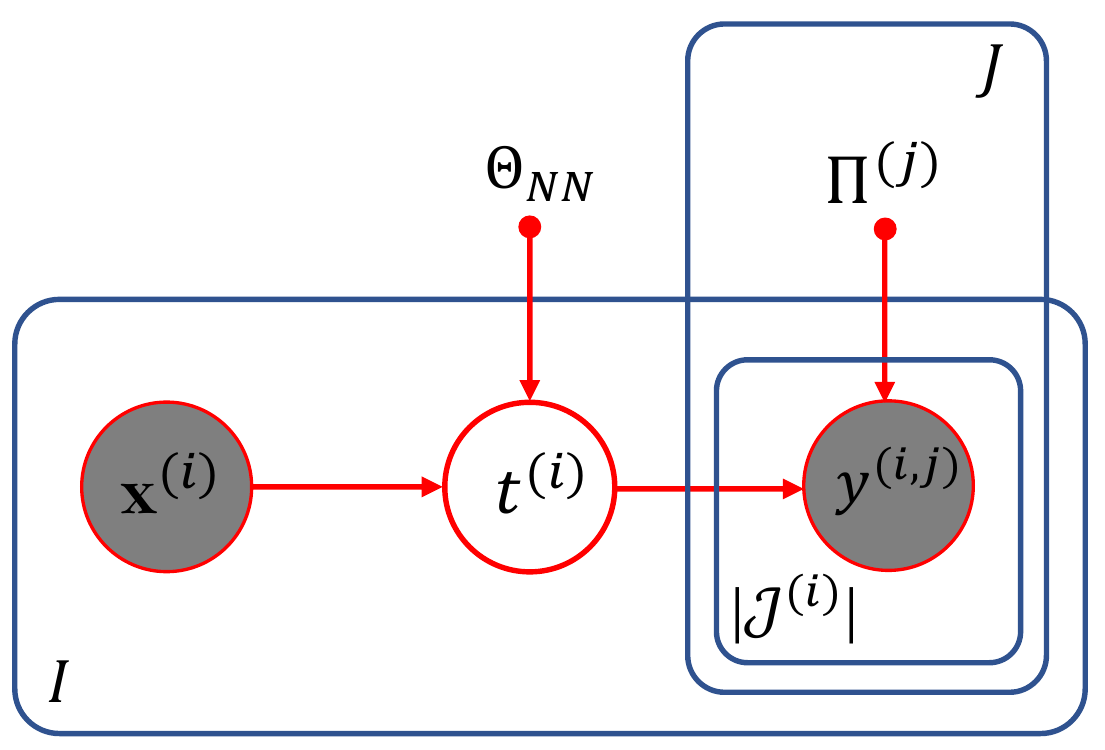} 
\caption{Probabilistic graphical representation of the basic model.}
\label{basic model}
\end{figure}

Let $\mathcal{D}=\{\mathbf{x}^{(i)}, \mathbf{y}^{(i)}\}_{i=1}^{I}$  be an i.i.d. dataset. For each instance
$\mathbf{x}^{(i)} \in \mathbb{R}^{D}$, we have a vector of noisy labels $\mathbf{y}^{(i)}=[y^{(i,1)},
\ldots,
y^{(i,J)}]$ given by a total of $J$ annotators, where 
$y^{(i,j)}$ represents the label by the  annotator  $j^{t h}$
on instance $\mathbf{x}^{(i)}$.
Each instance $\mathbf{x}^{(i)}$ has its corresponding  
\emph{unobserved}, latent ground truth $t^{(i)} \in\{1, \ldots, K\}$,
where $K$ represents the number of categories. 
In addition, considering that for each instance, often not every annotator has annotated it, we use $y^{(i, j)}=0$ to denote that annotator $j^{t h}$ has not annotated instance $\mathbf{x}^{(i)}$.
Our main goal is to train an accurate classifier for predicting
$t$ given new unknown instances $\mathbf{x}$ by using noisy training data $\{\mathbf{x}^{(i)}, \mathbf{y}^{(i)}\}_{i=1}^{I}$.

\subsection{Prior Art Utilized by Us}
\label{Prior Art Utilized by Us}
Our framework builds on a canonical  probabilistic graphical model~\cite{raykar2010learning,albarqouni2016aggnet} as a principled cornerstone. 
Here we introduce the basic model, whose graphical representation is shown in Figure~\ref{basic model}. 

In this latent variable model,
first, for each instance 
$\mathbf{x}^{(i)}$,
\begin{equation}
t^{(i)}|\mathbf{x}^{(i)} ; \Theta_{N N} \sim \operatorname{Cat}(t^{(i)} ; f_{\Theta_{N N}}(\mathbf{x}^{(i)}))\mathrm{,}
\label{eq:1}
\end{equation}
where the distribution of its unobserved ground truth  $t^{(i)}$
comes from a conditional categorical distribution $\operatorname{Cat}(t^{(i)} | f_{\Theta_{N N}}(\mathbf{x}^{(i)}))$; $f_{\Theta_{N N}}$ 
is a flexible neural network model parametrized by 
$\Theta_{N N}$.
Note that the classifier in the original work~\cite{raykar2010learning}  is a logistic regression.
Then, the reliability of the $j^{t h}$  annotator is modeled by an annotator-specific confusion matrix $\boldsymbol{\Pi}^{(j)}$, representing the likelihood of the annotator on identifying the ground truth label as any (other) labels. Formally, given $t^{(i)}$ and $\boldsymbol{\Pi}^{(j)}$, the distribution of crowd label
$y^{(i,j)}$ is determined by:
\begin{equation}
p(y^{(i, j)}=n | t^{(i)}=m ; \boldsymbol{\Pi}^{(j)})=\pi_{m n}^{(j)}\mathrm{,}
\label{eq:2}
\end{equation}
where $m, n \in\{1, \ldots, K\}$.
Based on the model construction, the optimization objective is to maximize the log conditional likelihood of the observed crowd labels given instance features, i.e., $\log p(\mathbf{Y} | \mathbf{X} ; \Theta)$,  with respect to the parameters
$\Theta=\{\Theta_{N N}, \boldsymbol{\Pi}^{(1)}, \ldots, \boldsymbol{\Pi}^{(J)}\}$. 
Those parameters can be learned by the EM algorithm.

\section{Method}
\label{method}

\begin{figure*}
    \centering
    \begin{minipage}{0.60\hsize}
        \includegraphics[width=\hsize]{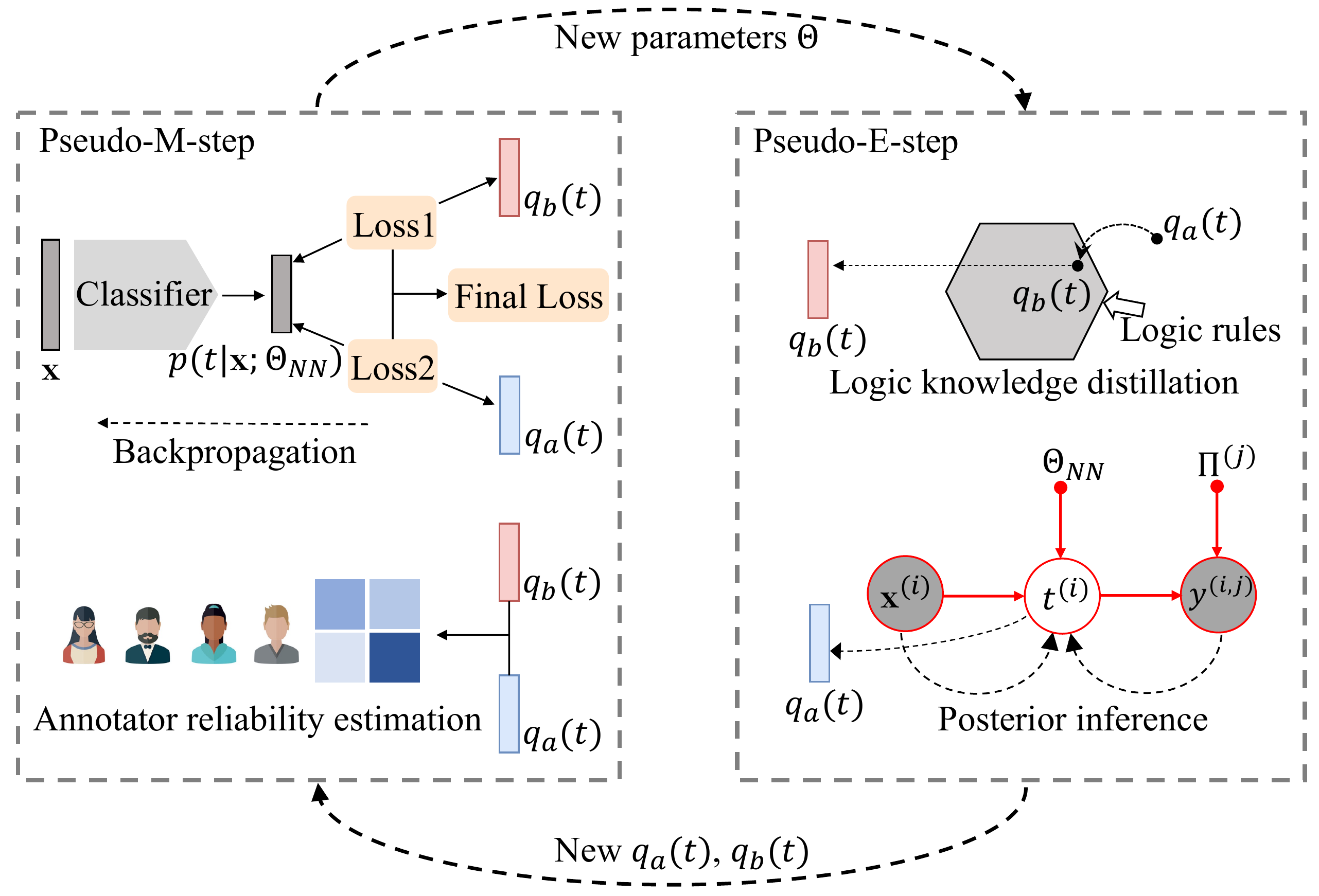}
    \end{minipage}
    \hfill
    \begin{minipage}{0.39\hsize}
        \begin{algorithm}[H]
\caption{Learning Logic-LNCL}
\label{alg:algorithm}
\textbf{Input}: 
Data $\mathcal{D}=\{\mathbf{x}^{(i)}, y^{(i)}\}_{i=1}^{I}$;
Rule set  
$\mathcal{R}=$ 
\quad
$ \left\{\left(R_{l}, w_{l}\right)\right\}_{l=1}^{L}$;
Parameter $k$:\ imitation strength;
Parameter $C$: regularization strength 
\\
\textbf{Output}: $\Theta_{N N}$ 
\begin{algorithmic}[1] %[1] enables line numbers
\STATE Initialize $q_{f}(t)$ with Majority Voting.
\WHILE{not converge}
\STATE \textbf{for} each minibatch in this epoch \textbf{do} \STATE  \quad 
update $\Theta_{N N}$ using backpropagation once
guided by
\textit{Loss}
 in Eq.~\ref{eq:8} or Eq.~\ref{eq:10}
(e.g., perform Eq.~\ref{eq:11} when using the mini-batch
gradient descent optimizer).
\STATE Update $\{\boldsymbol{\Pi}^{(1)}, \ldots \boldsymbol{\Pi}^{(J)}\}$  with Eq.~\ref{eq:12}.
\STATE
Update $q_{a}(t)$ with Eq.~\ref{eq:13}.
\quad 
\STATE Update $q_{b}(t)$  with Eq.~\ref{eq:15}.
\quad 
\STATE Update $q_{f}(t)$  with Eq.~\ref{eq:9}.
\ENDWHILE
\end{algorithmic}
        \end{algorithm}
    \end{minipage}
\caption{The EM-alike iterative procedure for learning parameters of Logic-LNCL. 
For all $I$ instances and $J$ annotators:
(\textit{i}) In the Pseudo-M-step, we update parameters of the  neural network and those of  $J$ annotators given two learning targets inferred from the Pseudo-E-step, i.e.,  truth posterior $q_{a}(t)$ (of each instance) obtained
from the conventional latent variable probabilistic model,
and 
$q_{b}(t)$ (of each instance) that contains the distilled logic knowledge;
(\textit{ii}) In the Pseudo-E-step, we infer the target $q_{a}(t)$ using the Bayes' theorem according to the
constructed
graph model, and infer the target $q_{b}(t)$ by adapting $q_{a}(t)$ with posterior regularization defined by the logic rules. 
}
\label{proposed_method}
\end{figure*}

This section introduces our proposed framework  Logic-LNCL depicted in Figure~\ref{proposed_method}.
We start by introducing the additional ``learning resources'', i.e. logic rules, in addition to the noisy labeled data mentioned in the Section~\ref{Problem Formulation: Learning from Noisy Data}.
Then we present the proposed EM-alike iterative learning algorithm.

\subsection{Logic Rules}
\label{logic rules}

% We encode first-order logic
% rules using probabilistic soft logic (PSL)~\cite{bach2017hinge}  for flexible encoding and stable optimization.

We introduce the off-the-shelf theory of probabilistic soft logic (PSL)~\cite{bach2017hinge}  to encode first-order logic rules.
Under the theory of PSL,
we  consider a set of 
first-order logic
rules with weights, i.e., 
$\mathcal{R}=\{(R_{l}, w_{l})\}_{l=1}^{L}$,
where $R_{l}$ is the $l$-th rule,
and $w_{l} \in[0, 1]$ 
represents the degree  of credibility or importance.
When all the variables in a logic rule are instantiated, it forms a “grounding”; and when the rules we defined are applied to data instances, a set of 
groundings of $R_{l}$,
denoted as 
$\left\{r_{l g}(\mathbf{x}, {t})\right\}_{g=1}^{G_{l}}$,
are formed. 
${G_{l}}$ refers to the total number of groundings instantiated in the  dataset for the $l$-th rule.
To illustrate rules in PSL, 
the following example encodes the voting behavior of two voters: 
\begin{equation}
\operatorname{friend}(B, A) \wedge \operatorname{votesFor}(A, P) \rightarrow \operatorname{votesFor}(B, P),
\end{equation}
where $\operatorname{friend}(B, A)$, $\operatorname{votesFor}(A, P)$ and 
$\operatorname{votesFor}(B, P)$
are called ``atoms'' (denoted as $\ell$), and their \textit{soft truth values} take value space of $[0,1]$ instead of the extremes $0$ and $1$ only.
PSL uses the following relaxation 
 of the logical conjunction ($\wedge$), disjunction ($\vee$), and
negation ($\neg$) for logic operators:
\begin{equation}
\begin{aligned}
I(\ell_{1} \tilde{\wedge} \ell_{2})
 &=\max \left\{0, I(\ell_{1})+I(\ell_{2})-1\right\}, \\
 I(\ell_{1} \tilde{\vee} \ell_{2})
 &=\min \left\{I(\ell_{1})+I(\ell_{2}), 1\right\}, \\
 I( \tilde{\neg}
l_{1})
 &=1-I(\ell_{1}),
\end{aligned}
\label{eq:4}
\end{equation}
where 
$I(\cdot)$ obtains the soft truth values.
In the above example, 
after instantiating
and $I(\operatorname{friend}(b, a)) = 1$ and  $I(\operatorname{votesFor}(a, p))= 0.9$, we have
$I(\text {friend}(b, a) \wedge \text {votesFor}(a, p)) = 0.9$.

As a remark, the logic rules introduced in our Logic-LNCL framework can be any first-order soft logic rules that follow the PSL formalism. Our framework is not compatible with other types of logic, such as the high-order logic.

\subsection{Pseudo-M-step: Learning}
\label{Pseudo-M-step: Learning}
In the Pseudo-M-step, we aim to use two learning targets to learn the parameters of the classifier and the annotators: the learning target from the conventional EM algorithm and a second target that contains the distilled logic knowledge.

Based on the probabilistic model and 
optimization objective $\log p(\mathbf{Y} | \mathbf{X} ; \Theta)$ constructed
in Section~\ref{Prior Art Utilized by Us}, the M-step in the conventional EM algorithm updates the parameters of the neural network
$\Theta_{N N}$ by maximizing\footnote{Derivation of the solution is given in Section~\ref{em_proof}.}:
\begin{equation}
\sum_{i=1}^{I} \text{num}(\mathcal{J}^{(i)}) \cdot \mathbb{E}_{q(t^{(i)})} \log[p(t^{(i)} \mid \mathbf{x}^{(i)} ; \Theta_{N N})],
\label{eq:5}
\end{equation}
or:
\begin{equation}
\sum_{i=1}^{I}  \mathbb{E}_{q(t^{(i)})}\log[p(t^{(i)} | \mathbf{x}^{(i)} ; \Theta_{N N})]\mathrm{,}
\label{eq:6}
\end{equation}
where $q(t^{(i)})$ is 
the ground truth estimation 
obtained from the previous E-step,
weight $\text{num}(\mathcal{J}^{(i)})$ is the number of annotators annotated
$\mathbf{x}^{(i)}$.
Equation~\ref{eq:6} comes from the strict derivation of the EM algorithm, and Equation~\ref{eq:5} comes from our adaptation of Equation~\ref{eq:6} with the addition of  weight $\text{num}(\mathcal{J}^{(i)})$.
The reason for considering the new objective function of Equation~\ref{eq:5} is that the 
instances with more annotations are likely to have more accurate posterior distribution.
Both of these  objectives allow the neural network to mimic the ground truth estimation $q(t^{(i)})$, and 
we have the flexibility to choose one of the objectives in our experiments.
In the following derivation, we assume that we adopt Equation~\ref{eq:6} for simplicity.

We did not straight follow Equation~\ref{eq:6} to optimize the classifier parameters.
Instead,
inspired by knowledge distillation~\cite{hinton2015distilling}, we introduce logic rules into the learning procedure by considering an extra learning target from   logic knowledge distillation, denoted as $q_{b}(t)$, in additional to the  original  learning target (Equation~\ref{eq:6}), denoted as $q_{a}(t)$. 
As shown in  Figure~\ref{proposed_method},
the loss function in our Pseudo-M-step is then formulated by:
\begin{equation}
\begin{aligned}
\sum_{i=1}^{I}\{-(1-k)\log \mathbb{E}_{q_{a}(t^{(i)})}[p(t^{(i)} | \mathbf{x}^{(i)} ; \Theta_{N N})] \\
+ k  \mathcal{L}(q_{b}(t^{(i)}), p(t^{(i)} | \mathbf{x}^{(i)} ; \Theta_{N N}))\}\mathrm{,}
\end{aligned}
\label{eq:7}
\end{equation}
where
$0<k<1$ is the imitation hyper-parameter balancing between the two parts;
$\mathcal{L}$ denotes
the cross-entropy loss function.
For the setting of $k$ here,
we can set it as a constant or a variable that constantly changes during the training progress.
The process of the neural network imitating $q_{b}(t)$ essentially transfers the logic knowledge into the neural network. 
Note that both the target $q_{a}(t)$ and $q_{b}(t)$ will be updated in the  Pseudo-E-step.
Considering the equivalence of negative log-likelihood and cross-entropy loss,
and the loss function in Equation~\ref{eq:7} is linear with respect to the label $t$,
the final loss function can be written as:
\begin{equation}
\begin{aligned}
\textit{ Loss }
:=
\sum_{i=1}^{I}  \mathcal{L}(q_{f}(t^{(i)}), p(t^{(i)} | \mathbf{x}^{(i)} ; \Theta_{N N}))\mathrm{,}
\end{aligned}
\label{eq:8}
\end{equation}
where 
\begin{equation}
q_{f}(t^{(i)})=(1-k) q_{a}(t^{(i)})+k q_{b}(t^{(i)}).
\label{eq:9}
\end{equation}
The weighted version of \textit{Loss} can be easily obtained as:
\begin{equation}
\sum_{i=1}^{I} \text{num}(\mathcal{J}^{(i)}) \cdot \mathcal{L}(q_{f}(t^{(i)}), p(t^{(i)} | \mathbf{x}^{(i)} ; \Theta_{N N})).
\label{eq:10}
\end{equation}
When we use standard mini-bath gradient descent to optimize the  
\textit{Loss}
in Equation~\ref{eq:8} (or Equation~\ref{eq:10}), the parameters will be updated:
\begin{equation}
\Theta_{N N} \leftarrow \Theta_{N N}-
\frac{\epsilon}{M}
 \nabla_{\Theta_{N N}} \sum_{\mathcal{D}_{M}} 
 \textit{Loss},
\label{eq:11}
\end{equation}
 where $\epsilon$ is learning rate, 
 $\mathcal{D}_{M}$ and $M$ are mini-batch data and batch size, respectively.

Both the original knowledge distillation method~\cite{hinton2015distilling}\footnote{
Hinton et al.~\cite{hinton2015distilling} found that a classifier can be better trained using the predictions of another well-trained classifier as the learning target together with the original training labels, compared to using the original labels only. 
Calling the procedure \textit{knowledge distillation}, they found there is a positive effect even when the classifier for providing help (called \textit{teacher}) is pre-trained with the same data as those later used for training the target classifier (called \textit{student}).}
and our framework forces the neural network (i.e.,  $p(t| \mathbf{x} ; \Theta_{N N})$) to imitate an additional  distilled knowledge  target in addition to a basic target. 
The major difference lies in that, the goal of Hinton et~al.~\cite{hinton2015distilling} is to better train a ``student'' neural network from another ``teacher'' neural network with better generalization ability, while our goal is to enable the neural network to learn from the  knowledge distilled from logic rules of interest; another technical difference is that, while the two learning targets in Hinton et~al.~\cite{hinton2015distilling} are fixed, our two targets and model parameters are
continuous iteratively updated during the EM-alike procedure.

We now proceed to updating the
% We also need to further optimize the 
parameters of the crowd annotators
$\{\boldsymbol{\Pi}^{(1)}, \ldots \boldsymbol{\Pi}^{(J)}\}$. 
Analogous to  optimizing the neural network parameters, we can optimize crowd annotators' parameters according to the probabilistic model constructed in Section~\ref{Prior Art Utilized by Us}
and 
follow the EM's
general recipe to directly obtain their closed-form solutions.
Here, we once again use $q_{f}(t)$ as the final ground truth estimation   to replace the original $q_{a}(t)$, and obtain\footnote{Derivation of the solution is given in Section~\ref{em_proof}.}:
\begin{equation}
\pi_{m n}^{(j)}=\frac{\sum_{i=1}^{I} q_{f}(t^{(i)}=m) \mathbb{I}(y^{(i, j)}=n)}{\sum_{i=1}^{I} q_{f}(t^{(i)}=m) \mathbb{I}(y^{(i, j)} \neq 0)}\mathrm{,}
\label{eq:12}
\end{equation}
where 
$y^{(i, j)} \neq 0$
means the annotator $j$ labeled the instance $i$;
and $\mathbb{I}(\cdot)$ is an indicator function that takes the value $1$ when 
the internal declaration  is true, and $0$ otherwise.

\subsection{Pseudo-E-step: Inference}
\label{E-step: Inference}

\subsubsection{Target $q_{a}(t)$ Construction}
Here, we assume that the variables
of interest strictly obey the basic
probabilistic graphical model (Figure~\ref{basic model}).
The posterior distribution of the latent variable $t$, $q_{a}(t)$, is inferred using the Bayes' theorem:
\begin{equation}
\begin{aligned}
q_{a}(t^{(i)}=k) \propto  p(t^{(i)}=k \mid \mathbf{x}^{(i)} ; \Theta_{N N})\cdot \\
\prod_{j \in \mathcal{J}^{(i)}} p(y^{(i, j)} \mid t^{(i)}=k ; \boldsymbol{\Pi}^{(j)})\mathrm{,}
\end{aligned}
\label{eq:13}
\end{equation}
where 
$\mathcal{J}^{(i)}$ represents all the annotators who annotated the $i^{t h}$ instance;
the parameters $\{\Theta_{N N}, \boldsymbol{\Pi}^{(1)}, \ldots, \boldsymbol{\boldsymbol{\Pi}}^{(J)}\}$ are from the previous Pseudo-M-step.

\subsubsection{Target $q_{b}(t)$ Construction}
\label{target qlt}
We aim to construct $q_{b}(t)$ such that it fits the pre-defined logic rules while making being as close to $q_{a}(t)$ as possible. To do so, we project $q_{a}(t)$ into a rule-regularized subspace using posterior regularization~\cite{zhu2014bayesian,hu2016harnessing}.

% We first impose the rule constraints on  $q_{b}(t)$ through an expectation operator:
We first use an expectation operator to 
impose the rule constraints on  $q_{b}(t)$:
$\mathbb{E}_{q_{b}(t^{(i)})}[v_{l}(\mathbf{x}^{(i)}, t^{(i)})]=1$,
where $v_{l}(\mathbf{x}^{(i)}, t^{(i)})$ is the rule's value of  $r_{l}(\mathbf{x}^{(i)}, t^{(i)})$\footnote{
$v_{l}(\cdot)$ corresponds to $1-d_{l}(\cdot)$ in probabilistic soft logic (PSL) formalism~\cite{kimmig2012short}, where $d_{l}(\cdot)$ is called as ``distance to satisfaction'' in PSL;
when a rule is fully satisfied, then $v_{l}(\cdot)=1$.
}. 
That is, we expect the distribution of $t$ in the grounded instance to conform to the logic rule. 
In the meanwhile, to keep $q_{b}(t)$ close to $q_{a}(t)$, we minimize the KL-divergence between those distributions. 
This leads to following optimization problem:
\begin{equation}
\begin{array}{c}
\min _{q_{b}(t^{(i)}), \xi \geq 0} \operatorname{KL}(q_{b}(t^{(i)}) \| q_{a}(t^{(i)}))+C \sum_{l} \xi_{l} \\
\text { s.t. } w_{l}(1-\mathbb{E}_{q_{b}(t^{(i)})}[v_{l}(\mathbf{x}^{(i)}, t^{(i)})]) \leq \xi_{l} \\
 l=1, \ldots, L,
\end{array}
\label{eq:14}
\end{equation}
where 
$C$ is the regularization parameter;
% $\xi_{l} \geq 0$ is the slack variable for the respective logic constraint. 
$\xi_{l} \geq 0$ denotes the slack variable of the respective logic constraint. 
This optimization problem, which is convex, can be solved through its dual form and obtain the closed-form solutions\footnote{Derivation of the solution is given in Section~\ref{Proof: Solving the Optimization Problem in Equation 12}.}:
\begin{equation}
q_{b}(t^{(i)}) \propto q_{a}(t^{(i)}) \exp \{-\sum_{l} C w_{l}[1-v_{l}(\mathbf{x}^{(i)}, t^{(i)})]\}
\label{eq:15}\mathrm{.}
\end{equation}
Intuitively, $v_{l}(\mathbf{x}^{(i)}, t^{(i)})$
will be greater for the cases when the corresponding rule is satisfied, thus increasing $q_{a}(t^{(i)}) \exp \{-\sum_{l} C w_{l}[1-v_{l}(\mathbf{x}^{(i)}, t^{(i)})]\}$, and vice versa.
From more intuitive perspective, the part $\exp \{-\sum_{l} C w_{l}[1-v_{l }(\mathbf{x}^{(i)}, t^{(i)})]\}$ 
 embedded with logic knowledge
plays the role of a ``constraint function'', or ``reward function''.

The overall learning procedure is summarized in Algorithm~\ref{alg:algorithm}.
We note that unlike all previous LNCL models, the proposed Logic-LNCL seamlessly integrates the following four processes---all necessary for allowing the classifier to learn simultaneously from noisy crowd labels and logic rules---in a principled EM-alike iterative logic knowledge distillation framework.
These processes are:
(\textit{i})
inference of latent ground truth;
(\textit{ii})
logic knowledge distillation attached to latent ground truth using   posterior regularization;
drawn on the knowledge distillation idea, 
(\textit{iii})
learning of classifier parameters, and 
(\textit{iv})
learning of annotator reliability.

\textbf{Implementation details: employ $q_{b}(t)$ at test phase.}
\label{Implementation details}
For instances without crowd labels at the test phase, we can employ either the trained neural network for prediction, i.e., $p(t| \mathbf{x} ; \Theta_{N N})$, or taking a step further, adapt this prediction using Equation~\ref{eq:15} to account for the logic rule. 
To do so, we simply need to replace  $q_{a}(t)$ by  $p(t| \mathbf{x} ; \Theta_{N N})$ in Equation~\ref{eq:15}.

\section{Applications}
\label{applications}
We instantiate our framework on two representative language tasks,  
text sentiment classification and named entity recognition (NER).
This section introduces the logic rules considered for integration.
Note that since NER is a typical class of sequence labeling tasks, it is also easy to extend the proposed framework to other parallel tasks, such as part-of-speech (POS) tagging.

\subsection{Sentiment Classification}
\label{Applications:Sentiment Classification}
In sentiment analysis, identifying contrastive sense in the sentence to capture the dominant sentiment precisely is a challenge for the classifier.
For humans, a useful clue of such contrastive sense is the widely-used conjunction word “but”, 
which is a vital sign for sentiment change within a sentence.
For those sentences containing “but”, the sentiment of clauses after “but” generally dominates.
Considering that a sentence \emph{S} has an “A-but-B” structure; we would expect the sentiment inclination of the entire sentence is consistent with the sentiment inclination of the  clause \emph{B}.
Formally, this can be expressed with the logic rule:
\begin{equation}
\text { positive (sentence \emph{S}) } \Rightarrow \boldsymbol{\sigma}_{\Theta}(\text {clause \emph{B}})_{+}\mathrm{,} 
\label{eq:16}
\end{equation}
\begin{equation}
\text { negative (sentence \emph{S}) } \Rightarrow \boldsymbol{\sigma}_{\Theta}(\text {clause \emph{B}})_{-}\mathrm{,} 
\label{eq:17}
\end{equation}
where
$\boldsymbol{\sigma}_{\Theta}(\text {clause \emph{B}})_{+}$
and 
$\boldsymbol{\sigma}_{\Theta}(\text {clause \emph{B}})_{-}$
denotes the probability
of clause 
$\emph{B}$
for ``positive and ``negative".
We set the weight of both rules to $1$.
Thus, we can get that the rule's value takes 
$\boldsymbol{\sigma}_{\Theta}(\text {clause B})_{+}$
when $y=+$ and $\boldsymbol{\sigma}_{\Theta}(\text {clause B})_{-}$ otherwise.

\definecolor{LightCyan}{rgb}{0.88,1,1}

\subsection{Named Entity Recognition}

For NER, the basic network largely ignore the constraint that should exist between valid consecutive labels. 
For instance, 
the labels before the ``I-ORG'' (``Inside-Organization'') can only be ``B-ORG'' (``Begin-Organization'') or ``I-ORG'' (``Inside-Organization'') but not ``B-PER'' (``Begin-Person'') etc.
Unlike recent work~\cite{lample2016neural} that adds a conditional random field (CRF) to
model bi-gram dependencies within label seqence, we apply the logic rules without introducing additional parameters that need to be learned.
Two example rules are:
\begin{equation}
\begin{array}{l}
\operatorname{equal}\left(t_{i}, I-ORG\right) \Rightarrow  \operatorname{equal}\left(t_{i-1}, B-ORG\right)\mathrm{,} 
\end{array}
\label{eq:18}
\end{equation}
\begin{equation}
\begin{array}{l}
\operatorname{equal}\left(t_{i}, I-ORG\right) \Rightarrow  \operatorname{equal}\left(t_{i-1}, I-ORG\right)\mathrm{.} 
\end{array}
\label{eq:19}
\end{equation}
The bi-gram dependencies are introduced by the logic rules that express the transition relationship which should exist within true labels.
With the theory of 
probabilistic
soft logic
and  the weights of the rules (e.g., $0.8$  and  $0.2$, which can be set through human knowledge, lightweight sample statistics, etc.), the weighted values of rules in Equations~\ref{eq:18}
and ~\ref{eq:19} 
can take  $0.8$  and  $0.2$, respectively.
According to the transition rules, we can use dynamic programming for efficient computation in Equation~\ref{eq:15}.

\label{Section:4}

\section{Proofs}
This section  provides two necessary proofs to support some of the conclusions in Section~\ref{method}. 
Section~\ref{em_proof} introduces parameter estimation of the basic graph model (to support Equations~\ref{eq:6},
\ref{eq:12}, and \ref{eq:13}), and Section~\ref{Proof: Solving the Optimization Problem in Equation 12}
introduces the derivation process for solving the
optimization problem in Equation~\ref{eq:14} (to support Equation~\ref{eq:15}).

\subsection{Proof: Parameter Estimation of The Basic Graph Model}
\label{em_proof}

For the basic graph model constructed  in Section~\ref{Prior Art Utilized by Us}, the log-likelihood
can be written as:
\begin{small}
\begin{equation}
\begin{aligned}
\log p(\mathbf{Y}\mid\mathbf{X} ; \Theta)=& \sum_{i=1}^{I}  \log p(\mathbf{y}^{(i)} \mid\mathbf{x}^{(i)} ; \Theta) \\
=& \sum_{i=1}^{I}  \log \sum_{t^{(i)}} p(\mathbf{y}^{(i)}, t^{(i)} \mid 
\mathbf{x}^{(i)};\Theta) \\
=& \sum_{i=1}^{I} \log \sum_{t^{(i)}} q_{i}(t^{(i)}) \frac{p(\mathbf{y}^{(i)}, t^{(i)} \mid \mathbf{x}^{(i)} ; \Theta)}{q_{i}(t^{(i)})} \\
\geq & \sum_{i=1}^{I}  \sum_{t^{(i)}} q_{i}(t^{(i)}) \log \frac{p(\mathbf{y}^{(i)}, t^{(i)} \mid \mathbf{x}^{(i)} ; \Theta)}{q_{i}(t^{(i)})}.
\end{aligned}
\tag{A.1}
\end{equation}
\end{small}
The last step of this derivation which obtains the 
$\sum_{i=1}^{I}  \sum_{t^{(i)}} q_{i}(t^{(i)}) \log \frac{p(\mathbf{y}^{(i)}, t^{(i)} \mid \mathbf{x}^{(i)} ; \Theta)}{q_{i}(t^{(i)})}$
(called ``\textit{Evidence Lower Bound} (ELBO)'') used Jensen’s inequality~\cite{bishop2006pattern}; for each $i$, $q_{i}(t^{(i)})$ can be some distribution over the $t^{(i)}$.
Note that we use $q(t^{(i)})$  to denote $q_{i}(t^{(i)})$  in Section~\ref{method} and  blow derivation for simplicity.

We then proceed to apply the
general
EM recipe to perform iterative calculations to solve the optimization problem:
\begin{equation}
\begin{aligned}
\text { E-step: Infer } q(t^{(i)}=k)& :=p(t^{(i)}=k \mid \mathbf{y}^{(i)}, \mathbf{x}^{(i)} ; \Theta) \\
\ \qquad & \propto p(t^{(i)}=k \mid \mathbf{x}^{(i)} ; \Theta_{NN})\cdot  \\
\qquad & \prod_{j \in \mathcal{J}^{(i)}} p(y^{(i, j)} \mid t^{(i)}=k ; \boldsymbol{\Pi}^{(j)}),
\end{aligned}
\tag{A.2}
\label{A.2}
\end{equation}
\begin{equation}
\begin{aligned}
&\text{ M-step: Learn }\\
\Theta &:=
\underset{\Theta}{\operatorname{argmax}}
\sum_{i=1}^{I}  \sum_{t^{(i)}} q(t^{(i)}) \log \frac{p(\mathbf{y}^{(i)}, t^{(i)} \mid \mathbf{x}^{(i)} ; \Theta)}{q(t^{(i)})} \\
&:=\underset{\Theta}{\operatorname{argmax}}\sum_{i=1}^{I} \mathbb{E}_{q(t^{(i)})} \log p(\mathbf{y}^{(i)}, t^{(i)} \mid \mathbf{x}^{(i)} ; \Theta) \\
&:=\underset{\Theta}{\operatorname{argmax}} \sum_{i=1}^{I}  [\mathbb{E}_{q(t^{(i)})} \operatorname{log} p(t^{(i)} \mid \mathbf{x}^{(i)} ; \Theta_{NN})+ \\
& \quad\qquad\quad\quad\quad\qquad\quad\mathbb{E}_{q(t^{(i)})} \log\prod_{j \in \mathcal{J}^{(i)}}p({y}^{(i,j)} \mid t^{(i)} ; \boldsymbol{\Pi}^{(j)})],
\end{aligned}
\tag{A.3}
\label{A.3}
\end{equation}
where
$\Theta = \{\Theta_{N N}, \boldsymbol{\Pi}^{(1)}, \ldots, \boldsymbol{\boldsymbol{\Pi}}^{(J)}\}$.
During the E-step, the posterior 
$q(t^{(i)})$
is obtained by using of Bayes's theorem given the parameters learned on the last M-step.
During the M-step, 
we can easily obtain the optimization objective in  Equation~\ref{eq:6}  according to Equation~\ref{A.3}, and we can obtain the closed-form solution of $\pi_{m n}^{(j)}$ 
by  equating the gradient of Equation~\ref{A.3} to zero.

\subsection{Proof: Solving the Optimization Problem in Equation~\ref{eq:14}}
\label{Proof: Solving the Optimization Problem in Equation 12}

Here we provide the derivation process for solving the optimization problem in Equation~\ref{eq:14} to obtain the solution in Equation~\ref{eq:15}. 
The optimization problem is:
\begin{equation}
\begin{array}{c}
\min _{q_{b}(t^{(i)}), \xi \geq 0} \operatorname{KL}(q_{b}(t^{(i)}) \| q_{a}(t^{(i)}))+C \sum_{l} \xi_{l} \\
\text { s.t. } w_{l}(1-\mathbb{E}_{q_{b}(t^{(i)})}[v_{l}(\mathbf{x}^{(i)}, t^{(i)})]) \leq \xi_{l} \\
 l=1, \ldots, L.
\end{array}
\tag{B.1}
\label{eq:B1}
\end{equation}
Here note that we assume that for instance $i$, each rule can act on it and form a grounding.

The following proofs are largely adapted from Ganchev et al.~\cite{ganchev2010posterior}
 and Hu et al.~\cite{hu2016harnessing}.
The Lagrangian of Equation~\ref{eq:B1} is:
\begin{equation}
\max _{\boldsymbol{\mu} \geq 0, \boldsymbol{\eta} \geq 0, \alpha \geq 0} \min _{q_{b}(t^{(i)}), \boldsymbol{\xi}} L,
\tag{B.2}
\label{eq:B2}
\end{equation}
where:
\begin{equation}
\begin{array}{l}
L=\mathrm{KL}(q_{b}(t^{(i)}) \| q_{a}(t^{(i)}))+\sum_{l}(C-\mu_{l}) \xi_{l} \\
\qquad+\sum_{l} \eta_{l}(\mathbb{E}_{q}\left[w_{l}(1-v_{l}(\mathbf{x}^{(i)}, t^{(i)}))]-\xi_{l}\right) \\
\qquad+\alpha(\sum_{t^{(i)}} q_{b}(t^{(i)})-1)\mathrm{.} 
\end{array}
\tag{B.3}
\label{eq:B3}
\end{equation}
Given Equation~\ref{eq:B2}, we can obtain:
\begin{equation}
\begin{array}{c}
\nabla_{q} L=\log q_{b}(t^{(i)})   +1-\log q_{a}(t^{(i)})
\\
+\sum_{l} \eta_{l}[w_{l}(1-v_{l}(\mathbf{x}^{(i)}, t^{(i)}))]+\alpha=0 \\
\Longrightarrow \quad q_{b}(t^{(i)})=\frac{q_{a}(t^{(i)}) \exp \{-\sum_{l} \eta_{l} w_{l}(1-v_{l}(\mathbf{x}^{(i)}, t^{(i)}))\}}{e \exp (\alpha)}\mathrm{;} 
\end{array}
\tag{B.4}
\label{eq:B4}
\end{equation}
\begin{equation}
\begin{aligned}
\nabla_{\xi_{l}} L=C-\mu_{l}-\eta_{l}=0  \\ \quad \Longrightarrow \quad \mu_{l}=C-\eta_{l}\mathrm{;} 
\end{aligned}
\tag{B.5}
\label{eq:B5}
\end{equation}

\begin{equation}
\begin{array}{c}
\nabla_{\alpha} L=\sum_{t^{(i)}} \frac{q_{a}(t^{(i)}) \exp \{-\sum_{l} \eta_{l} w_{l}(1-v_{l}(\mathbf{x}^{(i)}, t^{(i)}))\}}{e \exp (\alpha)}-1=0 \\
 \Longrightarrow  \alpha=\log (\frac{\sum_{t^{(i)}} q_{a}(t^{(i)}) \exp \{-\sum_{l} \eta_{l} w_{l}(1-v_{l}(\mathbf{x}^{(i)}, t^{(i)}))\}}{e})\mathrm{.} 
\end{array}
\tag{B.6}
\label{eq:B6}
\end{equation}
For simplicity, let 
$m_{t^{(i)}}=  \sum_{l} \eta_{l} w_{l}(1-v_{l}(\mathbf{x}^{(i)}, t^{(i)}))$,
$Z_{\eta}=\sum_{t^{(i)}} q_{a}(t^{(i)}) \exp \left\{-\sum_{l} \eta_{l} w_{l}\left(1-v_{l}(\mathbf{x}^{(i)}, t^{(i)})\right)\right\}$, $Z_{\eta, t^{(i)}}= q_{a}(t^{(i)}) \exp \left\{-\sum_{l} \eta_{l} w_{l}\left(1-v_{l}(\mathbf{x}^{(i)}, t^{(i)})\right)\right\}= \\
 q_{a}(t^{(i)}) \exp [m_{t^{(i)}}]$.
Plugging Equation~\ref{eq:B6} into Equation~\ref{eq:B4},
we obtain:
\begin{equation}
\begin{aligned}
\log q_{b}(t^{(i)}) 
&=\frac{\log Z_{\eta, t^{(i)}}}{\log Z_{\eta}} \\
&=\log q_{a}(t^{(i)}) 
+ m_{t^{(i)}}
- \log Z_{\eta}
\mathrm{,} 
\end{aligned}
\tag{B.7}
\label{eq:B7}
\end{equation}
\begin{equation}
\begin{aligned}
m_{t^{(i)}} &=
\log q_{b}(t^{(i)}) 
-
\log q_{a}(t^{(i)}) 
+
\log Z_{\eta} \mathrm{.} 
\end{aligned}
\tag{B.8}
\label{eq:B8}
\end{equation}
Plugging Equation~\ref{eq:B8} into $L$, we obtain:
\begin{equation}
\begin{aligned}
L &=-\log Z_{\eta}+\sum_{l}\left(C-\mu_{l}\right) \xi_{l}-\sum_{l} \eta_{l} \xi_{l} \\
&=-\log Z_{\eta}\mathrm{.} 
\end{aligned}
\tag{B.9}
\label{eq:B9}
\end{equation}
We have $\eta_{l} \leq C$ because of  Equation~\ref{eq:B5}, and since $Z_{\eta}$ monotonically decreases when $\eta$ increases, therefore:
\begin{equation}
\begin{aligned}
\max _{C \geq \eta \geq 0}-\log Z_{\eta} & \\
\Longrightarrow \quad & \eta_{l}^{*}=C\mathrm{.} 
\end{aligned}
\tag{B.10}
\label{eq:B10}
\end{equation} 
Plugging Equation~\ref{eq:B6}  and Equation~\ref{eq:B10}  into Equation~\ref{eq:B4}, we obtain the final solution of $q_{b}(t^{(i)})$ as in Equation~\ref{eq:15}.

\section{Evaluation}
\label{experiments}

\subsection{Setup}
\subsubsection{Datasets}
\label{datasets}
We evaluate our framework on sentiment classification and named entity recognition using two real-world public datasets\footnote{Code and datasets are available from~\url{https://github.com/junchenzhi/Logic-LNCL}.}: the Sentiment Polarity (MTurk) dataset~\cite{rodrigues2013learning} (balanced categories),
 and the CoNLL-2003 NER (MTurk) dataset~\cite{rodrigues2014sequence}. Crowd labels in both datasets were contributed by annotators from Amazon Mechanical Turk (MTurk).

\textbf{Sentiment Polarity (MTurk) Dataset.}
This crowdsourced dataset~\cite{rodrigues2013learning}  is constructed from the well-established
sentence-level
sentiment polarity dataset~\cite{pang-lee-2005-seeing}  with crowd labels.
The dataset contains sentences from movie reviews extracted from the website RottenTomatoes.com and the sentiment was classified as positive or negative.
The training data contains $4999$ sentences, $27747$ crowd labels annotated by $203$ different annotators.
The test data contains $5789$ sentences, which we shuffled and divided into a development set and a test set containing 3000 samples and 2789 samples, respectively.

\textbf{CoNLL-2003 NER (MTurk) Dataset.}
This crowdsourced dataset~\cite{rodrigues2014sequence}  is constructed from the well-established CoNLL-2003 NER dataset~\cite{sang2003introduction}  with crowd labels.
The goal is to recognize named entities (\textit{person, location, organization, miscellaneous}) together with their different parts (\textit{begin, inside}) in the sentence. 
All $9$ categories include: 
\textit{B-PER, I-PER, B-LOC, I-LOC, B-ORG, I-ORG, B-MISC, I-MISC, Others}.
The training set contains $5985$ sentences annotated by $47$ different annotators, whose F1 scores against the ground truth vary from $17.60 \%$ to $89.11 \%$.
We shuffled and divided the $3250$ test samples~\cite{rodrigues2014sequence} into a  development set and a test set containing 2000  samples and 1250 samples, respectively.
It is worth mentioning that,
crowd labels in the NER dataset may contain three main types of errors that annotators make during the annotation process:
(\textit{i})
\emph{Ignore error}: entity is not annotated by the annotator; 
(\textit{ii})
\emph{Boundary error}: entity type is correctly identified, but the annotated span does not match exactly; and 
(\textit{iii}) 
\emph{Span type error}: the entity type within the annotation span is wrong.

\subsubsection{Compared methods}
We consider the following representative LNCL (learning-from-noisy-crowds-labels) competitors:
(\textit{i}) MV-Classifier, a two-phase method that first estimates the ground truth from noisy labels by MV (Majority Voting), and then trains the classifier;
(\textit{ii}) GLAD-Classifier, similar to MV+Classifier, except the ground truth is estimated by GLAD \cite{whitehill2009whose};
(\textit{iii}) 
Raykar~\cite{raykar2010learning} and AggNet~\cite{rodrigues2013learning},
two representative probabilistic
models described in Section~\ref{Prior Art Utilized by Us};
(\textit{iv}) 
Three variants of the
state-of-the-art method CL~\cite{rodrigues2018deep} for deep learning from crowd labels: CL (MW), CL (VW), and CL (VW-B),
where ``MW'', ``VW'' and ``VW-B''
refer to  three different ways of parameterizing annotator reliability; 
(\textit{v}) 
 DL-DN$^{*}$ and DL-WDN$^{*}$~\cite{guan2018said},
which trains an individual network from the labels of each crowd annotator and aggregates the output from all networks for prediction;
% \jie{describe!}
(\textit{vi}) Our proposed variants Logic-LNCL-student and Logic-LNCL-teacher  using 
$p(t| \mathbf{x} ; \Theta_{N N})$,
and 
$q_{b}(t)$, as the output (details in 
paragraph
``Implementation details: employ $q_{b}(t)$ at test phase''), respectively;
Finally, we consider (\textit{vii}) Gold, denoting the classifier trained in the ideal case when true labels are known.

In addition, we analyze the 
\textit{inference}
 performance of the truth inference methods on the training set.
(1) On the sentiment polarity dataset, we consider the  baseline MV, the two state-of-the-art graph models DS~\cite{dawid1979maximum} and GLAD~\cite{whitehill2009whose}, and 
heuristic iterative inference based
PM~\cite{aydin2014crowdsourcing} and CATD~\cite{li2014confidence};
(2)
On the NER dataset, we consider the baseline MV, two 
NER-oriented extended graph models DS~\cite{dawid1979maximum} and IBCC~\cite{kim2012bayesian}  (which were originally designed for the traditional classification task), and
ad hoc NER-oriented graph models BSC-seq~\cite{simpson2018bayesian} and HMM-Crowd~\cite{nguyen2015deep}.

\subsubsection{Configurations}
\label{configurations}

\begin{figure}
\centering
\includegraphics[width=0.32\textwidth]{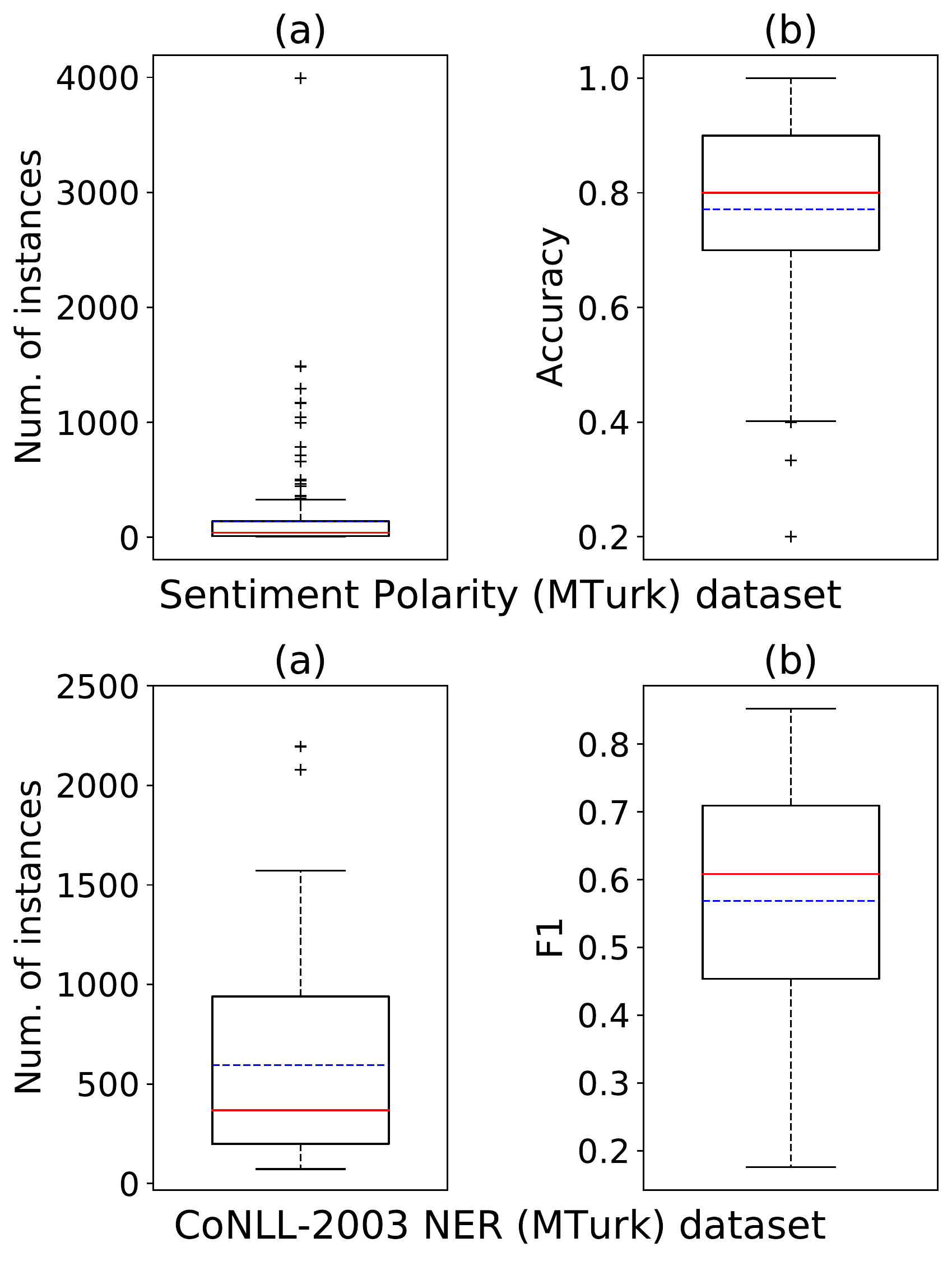} % Reduce the figure size so that it is slightly narrower than the column.
\caption{Boxplots for the number of
annotated instances (a)
and the accuracies/F1 (b) of the AMT annotators for
Sentiment Polarity (MTurk) dataset and 
 CoNLL-2003 NER (MTurk) dataset.}
\label{fig:4}
\end{figure}

\begin{table*}[t]
\centering
\begin{threeparttable}  
\begin{tabular}{l  l l l }
\toprule
& & \textbf{Sentiment  Dataset}
  & \textbf{NER Dataset }  
 \tabularnewline
\midrule
\multirow{7}{*}{Hyper-parameters}
& \texttt{Imitation strength 
$k^{(t)}$ ($t$: epoch)}
& 
$ \min \{1,1-0.94^{t}\}$ &  $\min \{0.8,1-0.90^{t}\}$  \\
& \texttt{Regularization strength $C$} & $5.0$  &  $5.0$  \\
& 
\texttt{Early stopping patience}
& $5$  & $5$  \\
& 
\texttt{Optimizer}
  &  Adadelta$^{\dag}$   & Adam$^{\S}$  \\
& 
\texttt{Learning rate}
 & $1.0$ + weight decay$^{*}$ & $0.001$$^{\S}$  \\
& 
\texttt{Epoch}
 & $30$$^{\dag}$ & $30$$^{\S}$  \\
% \noalign{\smallskip}
&
\texttt{Batch size}
 & 50$^{\dag}$    & 64$^{\S}$  
 \tabularnewline
\midrule
\multicolumn{2}{c}{Network architecture}
 &  Follow$^{\dag}$ & Follow$^{\S}$  
 \tabularnewline
\midrule
\multicolumn{2}{c}{Data division (train/development/test)}
   & 4999/3000/2789   & 5985/2000/1250
 \tabularnewline
\midrule
\multicolumn{2}{c}{Resource}    & \multicolumn{2}{c }{NVIDIA
Tesla V100 32GB GPUs}  
\tabularnewline
\bottomrule
\end{tabular}
\begin{tablenotes}
 \footnotesize   
\item[1] $^{\dag}$: The network architecture and  hyper-parameters 
are fully consistent with Kim ~\cite{kim-2014-convolutional}.
\item[2] $^{\S}$: 
The network architecture and  hyper-parameters are fully consistent with Rodrigues and Pereira ~\cite{rodrigues2018deep}.
\item[3] $^{*}$: 
For all methods on the sentiment dataset,
we added weight decay for original learning rate $1$ in Kim~\cite{kim-2014-convolutional}, i.e., decay by half every
$5$ epochs, to obtain more stable results.
\end{tablenotes}
\end{threeparttable} 
\caption{Experimental configurations.}
\smallskip
\label{Table:2}
\hfil
\captionsetup{type=table,skip=5pt}
\end{table*}

\begin{figure*}
\centering
\includegraphics[width=0.74\textwidth]{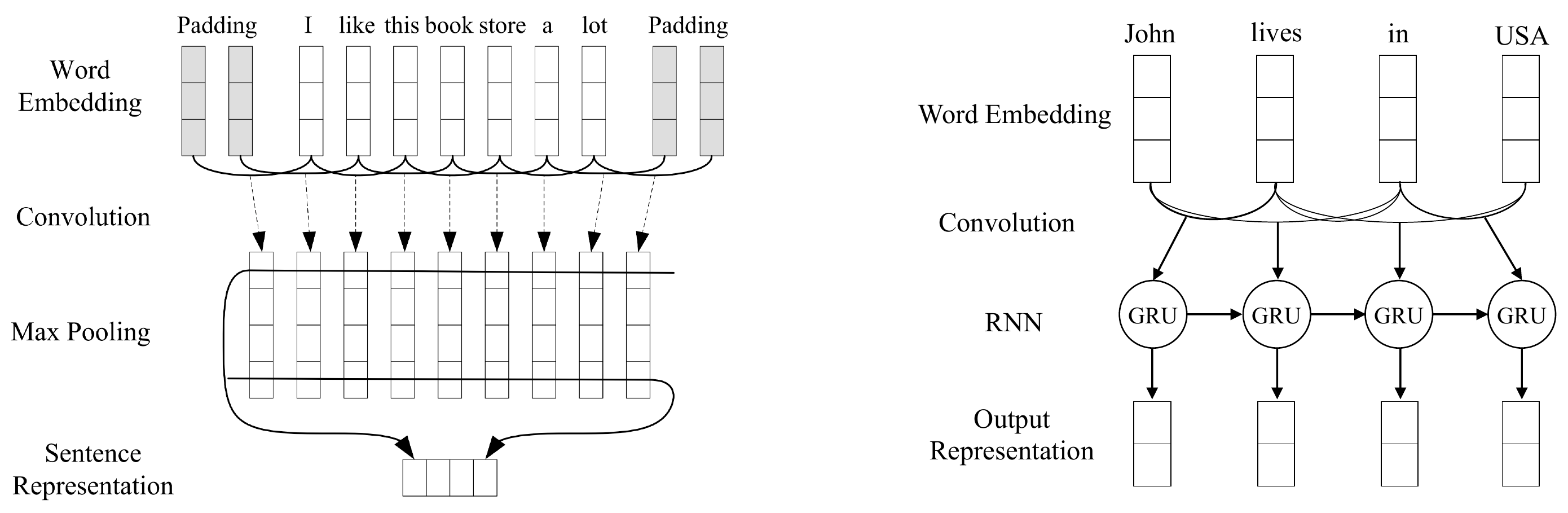} % Reduce the figure size so that it is slightly narrower than the column.
  \caption{{\textbf{Left}}: The  architecture for  sentiment polarity analysis introduced in Kim~\cite{kim-2014-convolutional}. 
It contains a convolutional layer (filter windows of 3, 4, 5 with 100 feature maps each, relu activation) on top of word vectors (300-demensionality ``static'' version bag-of-words embeddings), followed by a pooling layer and  a fully connected layer with softmax  activation (dropout of 0.5 on the penultimate layer with a constraint on l2-norms of 3).
  {\textbf{Right}}: 
  The architecture for NER
  introduced in Rodrigues and Pereira~\cite{rodrigues2018deep}.
It contains a layer of 300-dimensional GloVe word embeddings, followed by a 5x5 convolutional layer (relu activation) with 512 features, whose output (with 0.5 dropout) is then fed to a GRU cell (50 hidden states) and further fed to a fully-connected layer with a softmax output. }
\label{fig:base-nn}
\end{figure*}

% \begin{figure*}[t]
% \hspace{0.8in}
%   \subfigure  {\includegraphics[width=0.32\textwidth]{fig/cnn_sentiment.pdf}}
% \hspace{0.8in}
%   \subfigure
%   {\includegraphics[width=0.28\textwidth]{fig/classifier_ner-one.pdf}}
%   \caption{{\textbf{Left}}: The  architecture for  sentiment polarity analysis introduced in Kim~\cite{kim-2014-convolutional}. 
% It contains a convolutional layer (filter windows of 3, 4, 5 with 100 feature maps each, relu activation) on top of word vectors (300-demensionality ``static'' version bag-of-words embeddings), followed by a pooling layer and  a fully connected layer with softmax  activation (dropout of 0.5 on the penultimate layer with a constraint on l2-norms of 3).
%   {\textbf{Right}}: 
%   The architecture for NER
%   introduced in Rodrigues and Pereira~\cite{rodrigues2018deep}.
% It contains a layer of 300-dimensional GloVe word embeddings, followed by a 5x5 convolutional layer (relu activation) with 512 features, whose output (with 0.5 dropout) is then fed to a GRU cell (50 hidden states) and further fed to a fully-connected layer with a softmax output. }
% \label{fig:base-nn}
% \end{figure*}

\begin{table*}[t]
\centering
\begin{threeparttable}  
\begin{tabular}{l lcc >{\columncolor{lightgray!20}}c}
\toprule
\textbf{Paradigm}&
\textbf{Method}     & \textbf{Prediction}       &
\textbf{Inference$^{\dag}$  }
  & 
\textbf{ \underline{Average}}
  \tabularnewline
\midrule
\multirow{2}{*}{Two-stage LNCL} 
 &
MV-Classifier &  78.08    &  88.58  &  83.33 \tabularnewline
 &
GLAD-Classifier   & 78.45  & 
91.76
  &  85.11
\tabularnewline
\midrule
\multirow{7}{*}{One-stage LNCL} 
 &
 Raykar~\cite{raykar2010learning}   & - & 91.48 &  -
\tabularnewline
 &
 AggNet~\cite{albarqouni2016aggnet}   & 78.47
& 91.63 &  85.05
\tabularnewline
 &
 CL (VW)~\cite{rodrigues2018deep}   & 78.22      & 88.00   &  83.11
\tabularnewline
 &
 CL (VW-B)~\cite{rodrigues2018deep}   & 78.04      & 87.51  &  82.78
\tabularnewline
 &
 CL (MW)~\cite{rodrigues2018deep}   & 78.28    & 88.30 &  83.29
\tabularnewline
%\midrule
 &
\textbf{Logic-LNCL-student}     &   
78.85
 & \textbf{91.82}
 & 85.34
\tabularnewline
 &
\textbf{Logic-LNCL-teacher}      & \textbf{79.22}

&  \textbf{91.82}
 & \textbf{85.52}
\tabularnewline
\midrule
\multirow{5}{*}{Truth Inference} 
 &
MV      & -   &  88.58   &  -
\tabularnewline
 &
DS~\cite{dawid1979maximum}     & -    &  91.48  &  -
\tabularnewline
 &
GLAD~\cite{whitehill2009whose}     & -    &  91.76  &  -
\tabularnewline
 &
PM~\cite{aydin2014crowdsourcing}      & -   &  89.66 & -
\tabularnewline
 &
CATD~\cite{li2014confidence}     & -    &  91.49  &  -
\tabularnewline
\midrule
\multirow{1}{*}{-} &
\gc{Gold }      & \gc{79.26}    &  \gc{100}   &  \gc{86.13}  
 \tabularnewline
\bottomrule
\end{tabular}
\begin{tablenotes}
 \footnotesize   
\item[1] $^{\dag}$: Inference$^{\dag}$ denotes inference accuracy 
on the training set (results inferred by MV and GLAD; truth posterior estimation in 
% Raykar and 
AggNet; 
the classifiers' outputs in CLs;
$q_{f}(t)$ in Logic-LNCLs). 
\end{tablenotes}
\end{threeparttable} 
\caption{Performance (accuracy, $\%$) on the real-world Sentiment Polarity (MTurk)  dataset.
Results are averaged over 50 runs.}
\smallskip
\label{Table:3}
\hfil
\captionsetup{type=table,skip=5pt}
\end{table*}

\begin{table*}[t]
\centering
\begin{threeparttable}  
\begin{tabular}
{l lcccc cc
>{\columncolor{lightgray!20}}c}
% {lccccc}
 \toprule 
         \multicolumn{2}{c}{} & \multicolumn{3}{c}{\textbf{Prediction}} &
    \multicolumn{3}{c}{\textbf{Inference$^{\dag}$}} & \multicolumn{1}{c}{}  \\

  \cmidrule(lr){3-5} 
  \cmidrule(lr){6-8}

\textbf{Paradigm} &
\textbf{Method}
      & \textbf{Precision}
      
              & \textbf{Recall}
                        & \textbf{F1}
                            & \textbf{Precision}
      
              & \textbf{Recall}
                          & 
                          \textbf{ F1}
                          & \textbf{\underline{Avg. F1}}

\tabularnewline
% \midrule
% \midrule
\midrule
%  \hlinewd{0.75pt}

\multirow{1}{*}{Two-stage LNCL} &

 MV-Classifier & 65.14 &  45.98       & 53.89  &  79.12 &  58.50
 & 67.27   &  60.58
\tabularnewline

\midrule

\multirow{10}{*}{One-stage LNCL} & 
  AggNet~\cite{albarqouni2016aggnet}   &    61.67    &    58.64     & 60.09 &  77.19 &  73.02 & 75.04 &  67.57
\tabularnewline
& 
 CL (VW, 5$^{\S}$)~\cite{rodrigues2018deep}   & 69.37   & 52.11  & 59.32  &  79.19 &  71.72 & 75.25  &  67.29
\tabularnewline
%\midrule
& 
 CL (VW-B, 5$^{\S}$)~\cite{rodrigues2018deep}    & 58.23       & 59.92 &  58.97  &  75.27 &  73.41 & 74.30 &  66.64
\tabularnewline
%\midrule
& 
 CL (MW, 5$^{\S}$)~\cite{rodrigues2018deep} 
  & 62.98       &  61.57 & 62.19  & 78.37 &  75.14  & 76.70  &  69.45
  \tabularnewline
%\midrule
& 
CL (MW, 1$^{\S}$)~\cite{rodrigues2018deep}   &   53.75      &  44.70    & 48.19 &  61.93 &  50.21 &  54.42 &  51.31
\tabularnewline
& 
\textbf{Logic-LNCL-student}
     &  66.53    &  59.29     &  62.69
 &   84.90 &  74.11 & 
\textbf{79.14} &  70.92
\tabularnewline
& 
\textbf{Logic-LNCL-teacher}
      &  70.10      &   58.99 
  & \textbf{64.06}
 &  84.90 &  74.11
  & \textbf{79.14}
 &  \textbf{71.60} 
\tabularnewline
 \cmidrule(lr){2-9} 
  &
 CRF-MA$^{*}$~\cite{rodrigues2014sequence}    &     49.4    &     85.6   &    62.6 &  86.0 &  65.6  &    74.4   &  68.50
\tabularnewline
 &
 DL-DN$^{\ddagger}$~\cite{guan2018said}    &     72.3    &     45.9   &    56.2 &  - &  -  &    -   &  -
\tabularnewline
 &
DL-WDN$^{\ddagger}$~\cite{guan2018said}      & 61.1           &   48.0       & 53.4  &  - &  - &    - &  -
\tabularnewline
\midrule
\multirow{5}{*}{Truth Inference} & 
MV   &  -   &  -     &  -  & 79.12 &  58.50   & 67.27 &  -
\tabularnewline
 &
DS~\cite{dawid1979maximum}$^{\star}$  &  -   &  -     &  -  &  79.0 &   70.4  & 74.4 &  -
\tabularnewline
 &
IBCC~\cite{kim2012bayesian}$^{\star}$  &  -   &  -     &  -  &  79.0 &  70.4   & 74.4 &  -
\tabularnewline
 &
BSC-seq~\cite{simpson2018bayesian}$^{*}$   &  -   &   -      &  -  &  80.3 &  74.8   & 77.4 &  -
\tabularnewline
 &
HMM-Crowd~\cite{nguyen2017aggregating}$^{*}$   &  -   &  -     &  -  &  77.40 &  72.29   & 74.76 &  -
\tabularnewline
\midrule
-
& 
\gc{Gold (Upper Bound)}     &  \gc{72.52}   &   \gc{73.51}      &  \gc{72.98}  &  \gc{100} &  \gc{100}   & \gc{100} &  \gc{86.49} 
\tabularnewline
\bottomrule
\end{tabular}

\begin{tablenotes}
 \footnotesize   
\item[1] $^{\dag}$: Inference$^{\dag}$ refers to the meaning similar to the one in Table~\ref{Table:3}.
\item[2] $^{\S}$: $5^{\S}$ and $1^{\S}$
denote the number of times the classifier is pre-trained using the labels estimated from MV.
\item[3]  $^{*}$:
 results are reported from the original work.
 \item[4] $^{\ddagger}$ and $^{\star}$:
results are reported from 
Rodrigues and Pereira \cite{rodrigues2018deep},
Simpson and Gurevych~\cite{simpson2018bayesian},  respectively.
\end{tablenotes}
\end{threeparttable} 
\caption{Performance  ($\%$)  on the real-world 
 CoNLL-2003 NER (MTurk) dataset. Results  are averaged over 30 runs.
The best results under the F1 metric of most interest are marked in \textbf{bold}.}
\smallskip
\label{Table:4}
\hfil
\captionsetup{type=table,skip=5pt}
\end{table*}

We show the configurations in Table~\ref{Table:2}.
For the two datasets, 
the neural
network configurations in our proposed methods and the compared methods remain consistent with Kim~\cite{kim-2014-convolutional} and Rodrigues and Pereira~\cite{rodrigues2018deep}, respectively. 
Refer to Figure~\ref{fig:base-nn}
for more information about the networks.
For the objective function in our pseudo-M-step, we use the 
Equation~\ref{eq:6}
and 
Equation~\ref{eq:5}
on the two datasets, respectively.
As presented in Section~\ref{datasets} and on both datasets, we have separated a development set  from the original test samples to tune the hyper-parameters and to determine the  early stopping (epoch) time  for all methods.
Note that for the compared methods, there are no additional hyper-parameters.

\subsubsection{Evaluation Metrics}
\label{Evaluation Metrics}
For the sentiment polarity dataset containing two balanced categories, we use accuracy metrics to evaluate the performance.
For the NER dataset, we follow
prior work
~\cite{rodrigues2014sequence, nguyen2017aggregating, simpson2018bayesian, rodrigues2018deep} and use precision, recall and F1 under the \textit{strict criteria} rather than the other relaxed criteria for evaluation.

\subsection{Results and Analysis}
\label{Results and Analysis}
The results on the two datasets are shown in Table~\ref{Table:3} and Table~\ref{Table:4}, respectively.
In addition to the main results---the prediction performance and  the inference performance of the LNCL methods---we also show the inference performance of the ad hoc truth inference methods on the training set.
Here we first analyze the results of the most interest, i.e., the model's prediction performance.
We observe that both 
proposed 
Logic-LNCL-student and Logic-LNCL-teacher outperform all the compared methods, indicating that our approach has a higher level of robustness to label noise.
On the sentiment
polarity dataset, like the results reported in  Atarashi et al.~\cite{atarashi2018semi}, most methods exhibit similarly high accuracy. 
This is mainly because a relatively more number of annotations were received for each instance (5.55 on average), rendering the inference for true labels a relatively easy task for all comparison methods. Further, we utilize the \textit{t-test}  (unilateral statistics) to check whether the differences between our methods and the most competitive AggNet are statistically significant. In the two metrics of prediction and inference, when comparing Logic-LNCL-student and AggNet, the \textit{t-values} are $3.0$/$10.4$, and the \textit{p-values}  are both less than $0.01$ (i.e., \textit{statistically significant}); when comparing Logic-LNCL-teacher and AggNet, the \textit{t-values} are $5.7$/$10.4$, and the \textit{p-values} are both less than $0.01$.

\begin{figure*}[t] %这里使用的是强制位置，除非真的放不下，不然就是写在哪里图就放在哪里，不会乱动
	\centering  %图片全局居中
	\vspace{-0.0cm} %设置与上面正文的距离
	\subfigtopskip=-1pt %设置子图与上面正文或别的内容的距离
	\subfigbottomskip=0pt %设置第二行子图与第一行子图的距离，即下面的头与上面的脚的距离
	\subfigcapskip=-6pt %设置子图与子标题之间的距离
	\subfigure[Confusion matrix estimation]{
		\label{level.sub.1}
		\includegraphics[width=0.57\linewidth]{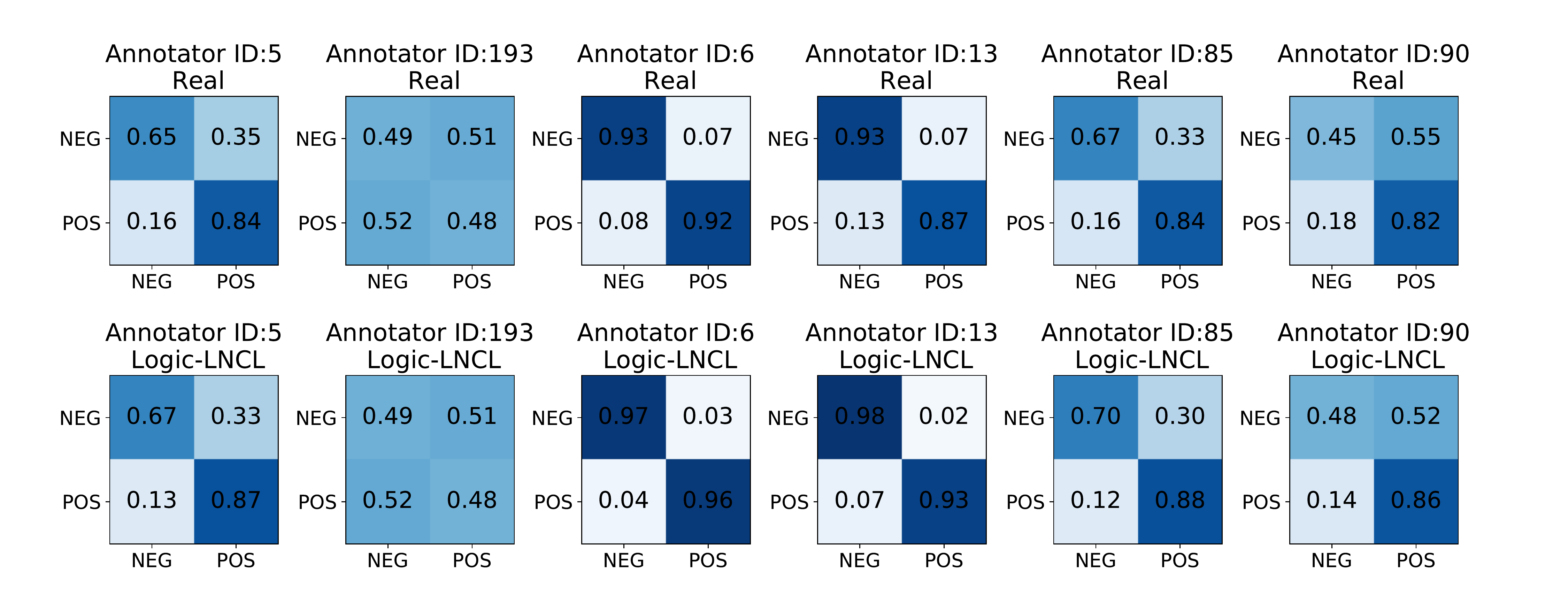}}
% 	\quad %默认情况下两个子图之间空的较少，使用这个命令加大宽度
	\subfigure[Overall reliability estimation]{
		\label{level.sub.2}
		\includegraphics[width=0.21\linewidth]{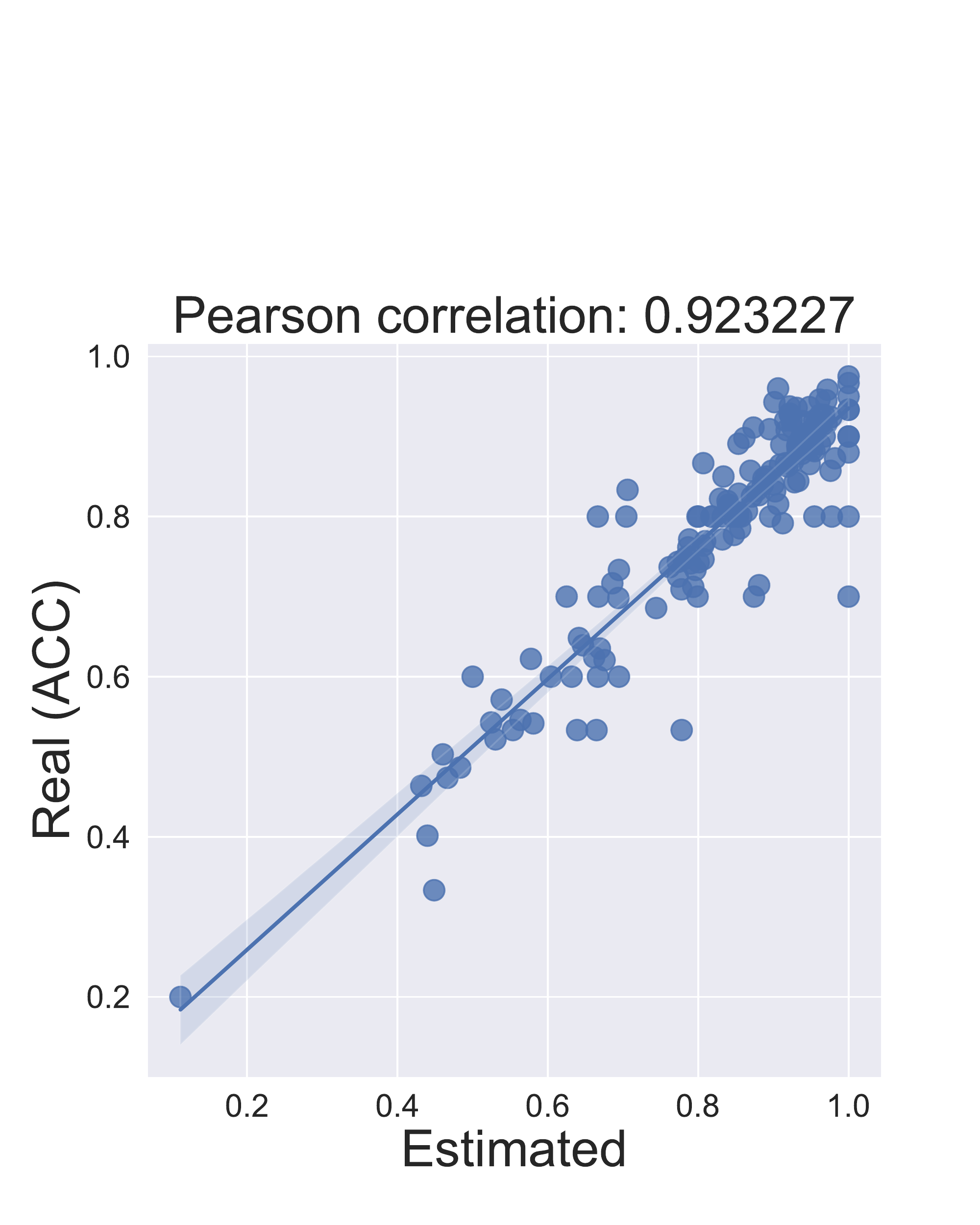}}
	  %这里是空了一行，能够实现强制将四张图分成两行两列显示，而不是放不下图了再换行，使用\\也行。
	\caption{Annotator reliability estimated by Logic-LNCL on the Sentiment Polarity (MTurk) dataset.
		}
	\label{Figure:6}
\end{figure*}

\begin{figure*}[t] %这里使用的是强制位置，除非真的放不下，不然就是写在哪里图就放在哪里，不会乱动
	\centering  %图片全局居中
	\vspace{-0.0cm} %设置与上面正文的距离
	\subfigtopskip=-1pt %设置子图与上面正文或别的内容的距离
	\subfigbottomskip=0pt %设置第二行子图与第一行子图的距离，即下面的头与上面的脚的距离
	\subfigcapskip=-6pt %设置子图与子标题之间的距离
	\subfigure[Confusion matrix estimation]{
		\label{level.sub.3}
		\includegraphics[width=0.69\linewidth]{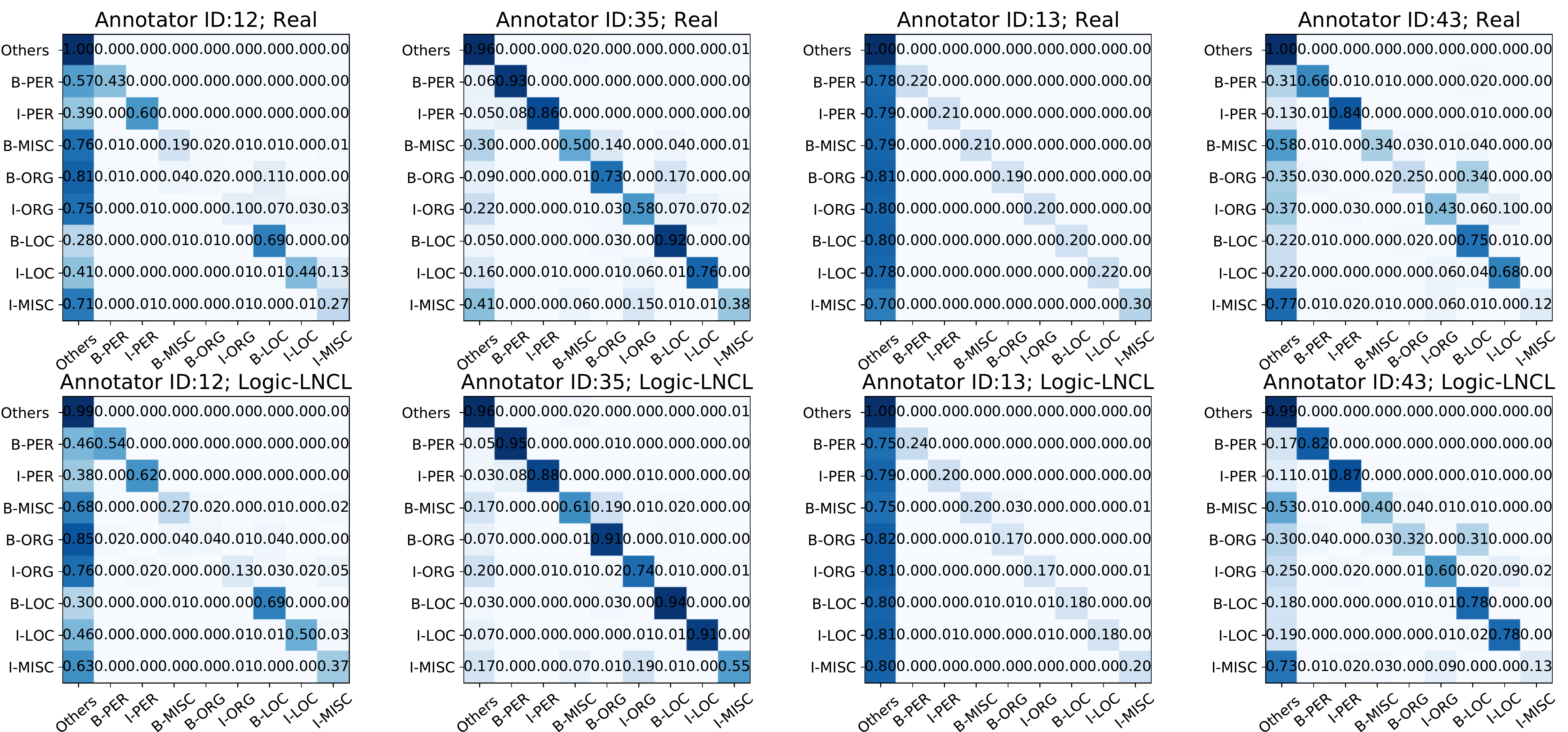}}
	\subfigure[Overall reliability estimation]{
		\label{level.sub.4}
		\includegraphics[width=0.21\linewidth]{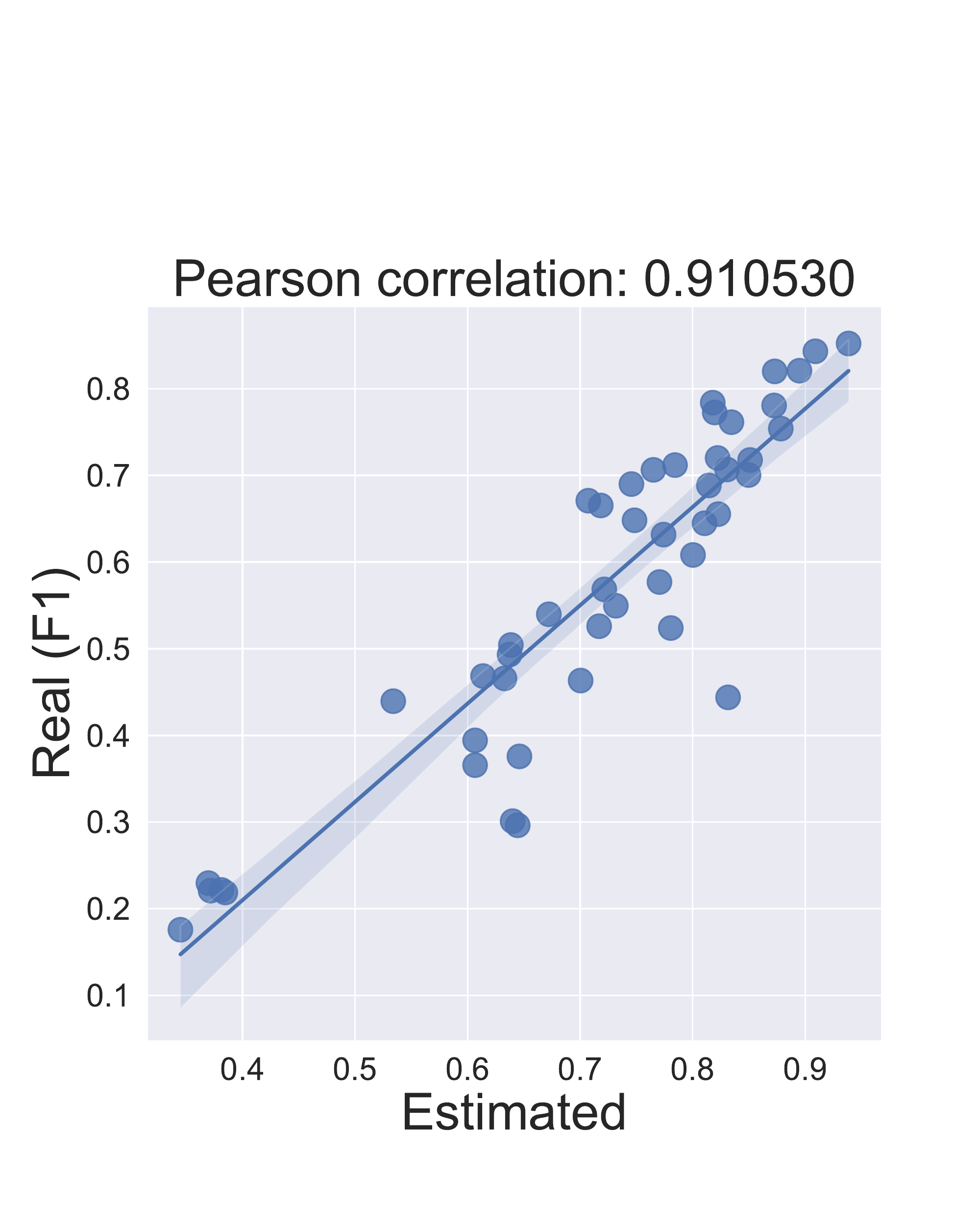}}
	\caption{Annotator reliability estimated by Logic-LNCL on the
	CoNLL-2003 NER (MTurk) dataset.
		}
	\label{Figure:7}
\end{figure*}

\begin{table*} 
\centering
% \resizebox{\linewidth}{14.2mm}
{
    \begin{tabular}{ l  c c    c c    >{\columncolor{lightgray!20}}c}
        \toprule 
         \multicolumn{1}{c}{} & \multicolumn{2}{c}{
         \textbf{Sentiment Dataset}
         }
         & \multicolumn{2}{c}{
         \textbf{NER Dataset}
         }
         & \multicolumn{1}{c}{}
         \\
         \cmidrule(lr){2-3} \cmidrule(lr){4-5}
        % \cmidrule(lr){8} 
  \textbf{Method}   
  &\textbf{Prediction} &\textbf{Inference} 
  &\textbf{Prediction} &\textbf{Inference} &\textbf{
  \underline{Average}
  }

  \\
 \midrule
  MV-Rule  &  78.41  & 88.96    & 47.66  & 61.63  & 69.17 
     \\
      GLAD-Rule  & 78.62 & 91.74   & 61.65  & 77.52   & 77.38
     % \\
     %   w/o-Rule  & 78.68  & 91.71  & 60.11  &  75.28 & 76.45
            \\
       w/o-Rule  & 78.47  & 91.63  & 60.11  &  75.28 & 76.37
     \\
MV-t  & 78.83  & 88.58   & 46.77 & 67.27 & 70.36  
     \\
our-other-rules-student
% {$\text{Logic-LNCL-student-bad}$}  
& 78.79  & 91.72   & 50.71 & 75.07 & 74.07
     \\
     our-other-rules-teacher  & 78.79  & 91.72   & 1.23 & 75.07 & 61.70  
     \\
\textbf{Logic-LNCL-student}
  & 78.85  & \textbf{91.82}    & 62.69  & \textbf{79.14}   & 78.13
     \\
     \textbf{Logic-LNCL-teacher}
  & \textbf{79.22} & \textbf{91.82}   & \textbf{64.06}  & \textbf{79.14}    & \textbf{78.56} 
     \\
    \bottomrule
    \end{tabular}
}
\caption{Results of ablation study.
Results (accuracy/F1, $\%$) on the two datasets are averaged over 50/30 runs.
(\textit{i}) MV-Rule and GLAD-Rule, where we ablate the
\emph{truth posterior inference}
  in Logic-LNCLs and add the
  \emph{same}
   logic rules as Logic-LNCLs', i.e., use the  labels estimated by MV/GLAD to replace  $q_{a}(t)$ to fed intro Equation~\ref{eq:15}; note that for NER, we use the labels estimated by AggNet to replace GLAD, which is inapplicable on NER;
(\textit{ii}) w/o-Rule, where we ablate the 
\emph{logic knowledge distillation}
  in Logic-LNCL;
(\textit{iii}) MV-t, where we add the logic knowledge for baseline MV-Classifier during the test phase, i.e., use the teacher model $q_{b}(t)$ to make predictions;
(\textit{iv})
our-other-rules-student and our-other-rules-teacher shows the results of our framework integrated with other rules: for sentiment classification, 
we replace the conjunction word ``but'' with the word ``however'' that has a relatively weaker indication for sentiment change;
for NER, we make an unrealistic assumption that 
each label type should be preceded by the same label type and without other possibilities, 
e.g., for the label ``I-ORG'', we only introduce the logic rule in Equation~\ref{eq:18} and ignore  Equation~\ref{eq:19}.
}%\smallskip
\label{Table:5}
\end{table*}

On the NER dataset, the performance of the compared methods shows a more considerable difference (Table~\ref{Table:4}). Logic-LNCL-teacher and Logic-LNCL-student achieve the best and second-best performance on the F1 metric, respectively. 
Both variants of our framework surpass the compared methods with a large margin: they outperform the state-of-the-art AggNet by 
$ 2.6 \%$ and $ 3.97 \%$, respectively, and 
vastly exceed
 the baseline MV-Classifier by $8.8\%$ and $10.17\%$, respectively.
An additional advantage of our methods compared to the competitive CL (MW) is that while CL (MW) relies on several epochs of pre-training on estimated labels with Majority Voting, 
our methods always learn from scratch 
without the need for additional pre-training on other methods.

\textbf{Employing $q_{b}(t)$ at test phase.} 
Comparing the two variants of Logic-LNCL, we observe that Logic-LNCL-teacher outperforms the Logic-LNCL-student on both datasets, with a more significant margin on the NER dataset. 
These results demonstrate the utility of logic
rules even in the test phase.

\textbf{Inference performance.}
\label{Inference performance}
 In addition to the above analysis of the generalization performance, we also observe in 
Table~\ref{Table:3} and Table~\ref{Table:4} 
 that the Logic-LNCLs outperform all compared methods (containing the ad hoc truth inference methods) in terms of inference performance on the training set.
Therefore, the logic knowledge distillation component in Logic-LNCLs can not only introduces ``dark knowledge''~\cite{hinton2015distilling} to the classifier to improve the generalization, but also 
\emph{can 
guide the ground truth inference process to directly improve inference accuracy}
 so as to feed the classifier with better training information.

\textbf{Advantage of sample-efficiency.}
Through further experiments, we find that on both datasets, both the
 student/teacher variants of our framework can achieve (and slightly exceed) the most competitive generalization performance in the compared methods (i.e., AggNet/CL(MW, 5)) using fewer training samples, i.e.,  $4300$/$3300$ out of the original $4999$ samples, 
$5700$/$4900$ out of the original $5985$ samples.
Additionally, under these sample conditions, our variants student/teacher also generally obtain the best inference performance on the training data of the two datasets: 
$91.78$/$91.72$, 
$78.92$/$78.58$.

\textbf{Annotator reliability estimation.}
 We further analyze the estimation performance of annotator confusion matrices 
$\{\boldsymbol{\Pi}^{j}\}_{j=1}^{J}$.
Figures~\ref{Figure:6} and ~\ref{Figure:7}
show the high accuracy of the annotator confusion matrices estimated with Logic-LNCL. Such a result further demonstrates the effectiveness of our method.
The way we obtained the four subfigures in Figures ~\ref{Figure:6} and ~\ref{Figure:7} is presented as follows.
In Figure~\ref{level.sub.1} and ~\ref{level.sub.3},
similar to Rodrigues et al.~\cite{rodrigues2018deep}, the six/four annotators with the largest number of labels were selected to analyze the confusion matrix  estimation.
We compare the estimated confusion matrices to the real ones, which are calculated based on their annotations and the ground truth. 
In Figure ~\ref{level.sub.2} and ~\ref{level.sub.4}, for each annotator, we divide the sum of the diagonal values of the estimated matrix by $2$ and $9$, respectively, to obtain a scalar to represent the overall reliability.
Figure~\ref{level.sub.2}  shows all annotators except the  anomalous annotators with five or fewer labels;
Figure~\ref{level.sub.4} shows all annotators.

\textbf{Ablation study.}
We compare with an extensive array of possible variants of our framework.
In Table~\ref{Table:5},
we observe:
(\textit{i})
Although MV-Rule and GLAD-Rule also distill knowledge from logic rules to provide the classifier to learn, they achieve suboptimal performance relative to Logic-LNCLs, mainly because they do not constantly refine the ground truth estimation during the iterative learning procedure like ours;
(\textit{ii})
On both datasets, our student model and teacher model outperform the important variant w/o-Rule. In particular,  on the NER dataset, our models outperform w/o-Rule in prediction performance by 
$ 2.58 \%$ and $ 3.95 \%$, respectively;
(\textit{iii})
The results of variants concerning relatively bad rules corroborate our intuition that  rules that do not conform well to objective laws can not make the framework work satisfactorily;
(\textit{iv})
Our Logic-LNCLs surpasses all other competitors, 
\emph{demonstrating the superiority of  seamlessly integrating the four essential processes in the principled iterative distillation framework: inference of ground truth, knowledge distillation from logic rules, learning of neural network and learning of annotator reliability; the interplay between these  processes makes them mutually beneficial.}

\section{Related Work}
\label{relatedwork}

\textbf{Learning from noisy crowd labels}
has long been the goal of the
weakly supervised learning
community~\cite{zhou2018brief}. 
Back in early 1970s,
Dawid and Skene~\cite{dawid1979maximum}  proposed a probabilistic graph model to inference the latent ground truth labels using the expectation maximization (EM) algorithm. 
This truth inference process is followed by a traditional supervised learning phase to  complete the learning~\cite{zheng2017truth}. 
Since then, a variety of generalizations and improvements upon the original method have been proposed under the same probabilistic framework~\cite{whitehill2009whose,welinder2010multidimensional,liu2012variational,han2016incorporating,tian2018max}.
Another line of the research~\cite{raykar2010learning,rodrigues2013learning,albarqouni2016aggnet,atarashi2018semi,rodrigues2018deep,li2020coupled,chen2020structured,li2021learning} investigates learning directly from noisy crowdsourced labels without the preceding truth inference stage. 
To the best of our knowledge, our work is the first that considers the integration of symbolic logic knowledge for learning from noisy crowd labels.

\textbf{Learning from noisy labels}
 is a closely related, larger research area that concerns the presence of noisy labels, regardless of the sources of such labels. In this learning setting, the labels are not necessarily from crowds and often the case, there is only one noisy label for each instance. 
A large bulk of work has been proposed investigating different learning strategies, e.g., designing more robust loss function~\cite{zhang2018generalized,patrini2017making},  architecture~\cite{sukhbaatar2014training,chen2015webly},  regularization~\cite{goodfellow2014explaining,pereyra2017regularizing}, or a better way of selecting samples~\cite{malach2017decoupling}, etc. 
Essentially, many robust architecture-based methods in this learning setting~\cite{sukhbaatar2014training,chen2015webly} are similar to those for learning from noisy crowd labels~\cite{raykar2010learning,albarqouni2016aggnet,rodrigues2018deep,chen2020structured}: they all infer the ground truth labels through a neural layer from the noisy labels.

\section{Discussion}
\label{discussion}

We recently noticed that there had been related work~\cite{zhang2021wrench} using LNCL methods to perform the similar \textit{weak supervision} learning problem~\cite{zhang2021wrench, ratner2017snorkel}, where annotation sources come from multiple programs defined artificially in advance instead of multiple crowd annotators. We plan to deploy our framework on weak supervision datasets in the future to demonstrate its  versatility further. 
\textit{At a higher level, we advocate that for all learning tasks with ground truth  latent variables (e.g., the extensively studied leaning-from-noisy-labels  task presented in Section~\ref{relatedwork}), 
the idea of Logic-LNCL can be transferable for iterative computation of learning and inference, where the inference process seamlessly combines the truth inference and  logic knowledge distillation.}
Hence, we  plan to extend our framework to the  \textit{learning-from-noisy-labels} task and conduct further empirical validation.
For the limitations, one of the most noteworthy points we believe is that for vision tasks other than language tasks, it is more  difficult to extend our framework yet.
Because we need to propose a more complex unified framework containing the indispensable process of embedding continuous high-dimensional data into the semantic space to represent high-level concepts (used for logics). 

\section{Conclusion and Future Work}
\label{conclusion}

This paper proposes to incorporate symbolic logic knowledge into deep learning from noisy crowd labels. We introduce Logic-LNCL, an EM-alike iterative logic  knowledge distillation framework that integrates logic knowledge with the inference for true labels, allowing the  neural network to learn from both the logic knowledge and the data. 
Our method, therefore, benefits both from the flexibility of deep learning and the robustness of logic rules. 
Our experiments demonstrate that Logic-LNCL significantly improves base networks on sentiment classification and named entity recognition with logic rules. 
The encouraging results show a strong potential of our proposed framework for improving other learning tasks, such as vision tasks, which we plan to study in the future.

\section*{Acknowledgment}
This work was supported by National Natural Science Foundation of China Under Grant Nos (61932007, 61972013, 62141209).

\bibliographystyle{IEEEtran.bst}
\bibliography{example_paper}
\newpage

\end{document}